\documentclass[sigconf,authorversion,noacm]{acmart}
\renewcommand\footnotetextcopyrightpermission[1]{} 
\AtBeginDocument{%
  \providecommand\BibTeX{{%
    \normalfont B\kern-0.5em{\scshape i\kern-0.25em b}\kern-0.8em\TeX}}}

\copyrightyear{2020} 
\acmYear{2020} 
\setcopyright{acmcopyright}
\acmConference[MM '20]{Proceedings of the 28th ACM International Conference on Multimedia}{October 12--16, 2020}{Seattle, WA, USA}
\acmBooktitle{Proceedings of the 28th ACM International Conference on Multimedia (MM '20), October 12--16, 2020, Seattle, WA, USA}
\acmPrice{15.00}
\acmDOI{10.1145/3394171.3413612}
\acmISBN{978-1-4503-7988-5/20/10}

\begin{document}
\fancyhead{}

\title{Learnable Optimal Sequential Grouping for Video Scene Detection}

\author{Daniel Rotman, Yevgeny Yaroker, Elad Amrani, Udi Barzelay, Rami Ben-Ari}
\email{[danieln@il.,yevgenyy@il.,elad.amrani@,udib@il.,ramib@il.]ibm.com}
\affiliation{
  \institution{IBM Research}
  \city{Haifa}
  \country{Israel}
}


\begin{abstract}
Video scene detection is the task of dividing videos into temporal semantic chapters.
This is an important preliminary step before attempting to analyze heterogeneous video content.
Recently, Optimal Sequential Grouping (OSG) was proposed as a powerful unsupervised solution to solve a formulation of the video scene detection problem.
In this work, we extend the capabilities of OSG to the learning regime.
By giving the capability to both learn from examples and leverage a robust optimization formulation, we can boost performance and enhance the versatility of the technology.
We present a comprehensive analysis of incorporating OSG into deep learning neural networks under various configurations.
These configurations include learning an embedding in a straight-forward manner, a tailored loss designed to guide the solution of OSG, and an integrated model where the learning is performed through the OSG pipeline.
With thorough evaluation and analysis, we assess the benefits and behavior of the various configurations, and show that our learnable OSG approach exhibits desirable behavior and enhanced performance compared to the state of the art.
\end{abstract}


\settopmatter{printacmref=false}

\maketitle

\section{Introduction}
\label{sec:intro}

With video content rapidly growing in quantity and availability, it becomes crucial to develop the relevant technologies to analyze, classify, and understand the content in videos.
However, one of the biggest issues when dealing with videos is analyzing the temporal aspect.
When dealing with heterogeneous video content, it is crucial to be able to partition a video into semantic scenes before performing any sort of algorithmic analysis.
Besides contextual analysis, division to scenes can facilitate automatic construction of a table of contents, video summarization, chapter skimming, and more \cite{intro_tasks_table_of_contents,intro_tasks_summarization,intro_tasks_browsing,intro_tasks_analysis}.

Video scenes are an ingrained part of the hierarchical structure of videos.
At the finest level of division, a video is composed of a series of images called \textit{frames}.
A sequence of frames captured from the same camera at the same time is called a \textit{shot}.
Identifying the shot transitions is considered somewhat a solved problem due to the relative uniformity of the frames in a shot \cite{intro_sbd}, and established methods can be used off-the-shelf with impressive performance \cite{sbd,sbd2}.
A group of shots relating a specific event or narrative is called a \textit{scene}.
A formal definition of a scene is given by \cite{scene_definition}, as a sequence of semantically related and temporally adjacent shots depicting a high-level concept or story.
Identifying the transition locations between scenes is considered a much higher-level problem, and which is the focus of this work.

Recently, we proposed Optimal Sequential Grouping (OSG) \cite{ours_ism} as an effective deterministic optimization formulation to solve the video scene detection problem.
The approach takes the distance matrix of the shot representations and calculates the optimal division given a cost function on the intra-scene distances \cite{ours_icmr}.
Despite its generality and strengths, the formulation leaves no room for learning from examples.

For learning from examples, deep learning has risen in popularity in recent years as a leading technology in many fields, and doubly so in the field of computer vision \cite{intro_deeplearning_computervision}.
However, it can be beneficial to combine learning with analytical deterministic algorithms to gain the advantages of both learning from examples and incorporating designer knowledge and expertise \cite{intro_deeplearning_analytical1,intro_deeplearning_analytical2,intro_deeplearning_analytical3,intro_deeplearning_analytical4}.
Merging learning with deterministic formulations can help arrive at more explainable technologies, ensure validity of performance, guide parameter learning, and support generalization as opposed to memorising.

Therefore, in this work, we present an approach to integrate the OSG formulation into a deep learning setting.
Our model retains the original strengths of attaining an optimal division given the defined cost function, but additionally has the ability to learn better representations given annotated scene divisions.

\begin{figure*}[t]
\centering
\includegraphics[width=\linewidth]{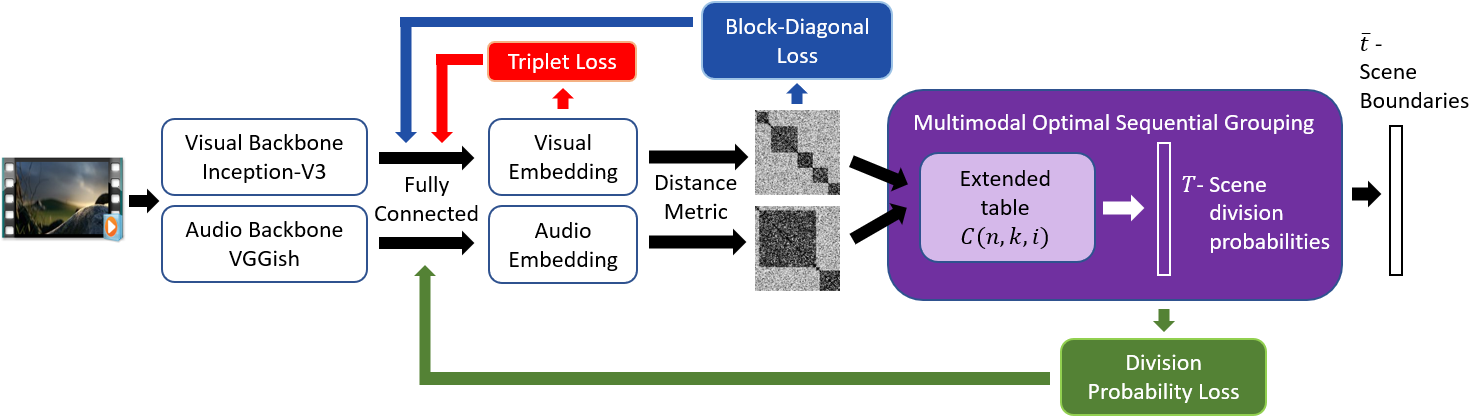}
   \caption{Our three configurations of OSG incorporated into a neural network: 1) the triplet loss (in red, OSG-Triplet), or 2) block-diagonal loss (in blue, OSG-Block), or 3) scene division probability loss (in green, OSG-Prob). The latter includes learning through the pipeline of the OSG dynamic programming algorithm, and aggregating the probabilities for division at specific locations. Video frame © Blender Foundation | gooseberry.blender.org.}
\label{fig:model}
\end{figure*}

We present a number of possible configurations for integrating OSG into the learning regime with different levels of integration.
First, we present the use of the triplet loss \cite{facenet}, for a classical learning approach to train an embedding with valuable properties for division into scenes.
This embodies the most straightforward and logical approach, but does not incorporate directly the properties of OSG.
Next, we present a tailored loss directly on the distance matrix values.
This loss is aimed to provide the input data in a representation which is favorable for the OSG formulation.
Finally, we present an approach where learning is performed through the OSG pipeline and dynamic programming formulation to allow direct learning from results.

Figure \ref{fig:model} depicts our OSG model with the possible configurations.
We analyze the performance and results of the different approaches and configurations.
Besides out-performing the state of the art, we show how the different configurations function and analyze the behavior, benefits, and advantages.

\section{Previous Work}
\label{sec:previous}

In this section we review some of the recent work on video scene detection where the task is focused on creating a complete partitioning of a motion-picture film using visual features.
For a more complete review including, for example, transcript-based approaches, news segmentation, and scene retrieval, see \cite{vsd_survey}.

\subsection{Video Scene Detection}
\label{sec:previous_vsd}

\subsubsection{Unsupervised Approaches.}

The prior art of video scene detection consists of mostly unsupervised approaches even in the most recent works.

A prevalent approach for scene detection is to perform a variety of clustering techniques \cite{lb,lb_measure,prior_cluster}. By representing video shots in some feature space the assumption is that shots from the same scene will cluster together. The weakness with such an approach is that the temporal aspect is not an inherent part of the formulation and is usually either enforced by post-processing or integrated into the feature space (as weighting, or as an additional dimension) instead of being an integral aspect of the problem.

Graph approaches \cite{prior_graph1,prior_graph4,prior_vsd_2_annotation} denote shots as nodes in a graph and perform graph analysis algorithms to determine the scene transitions using the graph cut algorithm. Additionally, \cite{prior_stg,mklab} construct Scene Transition Graphs by representing clusters of shots as nodes and calculating a cumulative confidence for the locations of scene divisions leveraging the primary set algorithm.

Regarding other advanced methods, \cite{prior_sequence_alignment} perform sequence alignment on shots categorized by clustering to identify recurring themes and production rules.
\cite{prior_vsd_1_bow} group shots with a bag of visual words descriptor and perform a sliding window for combining shots or short scenes together. \cite{prior_dynamic_programming} perform dynamic programming with a heuristic search scheme of boundaries calculated by linear discriminant analysis over shot similarities.

A specific brand of video scene detection which is of high interest focuses on egocentric videos \cite{prior_ego1,prior_ego2,prior_ego3}. 
Despite the overlap, the type of challenges and the level of variability in a scene when captured by an egocentric camera are not comparable to the complexity of a movie scene which demands a higher level of semantic understanding.

\subsubsection{Deep Learning for Video Scene Detection.}

Regarding methods for video scene detection which incorporate deep learning, one less recent but relevant method \cite{prior_lb_siam} performs learning a distance measure using a deep siamese network and applies spectral clustering to approximate scene boundaries.
They learn a joint representation of visual features and textual features (obtained from video transcription) for a similarity metric to represent the video.
This method has the most in common with our approach.
The main differences are that the authors do not incorporate the learning pipeline into the scene division.
The learning is performed to train a distance metric, but not tailored to the spectral clustering segmentation or learned backward directly from the segmentation results.
Therefore the learning could be seen as detached from the division stage, similar to the triplet configuration we present below (see Section \ref{sec:triplet}).

In \cite{prior_vsd_3_lstm}, the authors use deep visual, CSIFT, and MFCC audio features to represent shots.
They apply a CNN architecture to each input modality  and train an LSTM model to output whether each shot is a scene transition or not.
This is one of the most advanced approaches with learning video scene detection in a straight-forward manner.
However, results are comparable to \cite{prior_lb_siam} which means there is most likely room for improvement.

\subsection{Optimal Sequential Grouping}
\label{sec:previous_osg}

In this work, we focus on our recently proposed method for video scene detection: Optimal Sequential Grouping (OSG).

In \cite{ours_ism}, we presented OSG as a dynamic programming algorithm to divide a video by finding the optimal solution to an additive cost function.
The cost function sums the \textit{block-diagonal} of the distance matrix which represents the intra-scene distances between shots.
We additionally presented a \textit{log-elbow} method to estimate the number of scenes directly from the distance matrix.
We extended this work \cite{ours_mmsp} to utilize multiple modalities in the OSG framework.
By using an intermediate fusion approach, we merged the separate sequential grouping divisions into a single decision.
In \cite{ours_icmr}, we presented a new normalized cost function with analytically superior mathematical properties, and used deep features to represent the videos.
Despite the beneficial mathematical properties, we chose to forgo using the normalized cost function in this work, because the normalization adds an additional computation complexity to the dynamic programming solution. We believe that when incorporating learning into OSG (as we detail in this paper), the resulting distance values will likely overcome the mathematical bias, making the choice of cost function less critical.

The technical details of the formulation and solution of OSG are expanded on in Section \ref{sec:osg}.

Our contributions are as follows:
(1) We are one of the first to explore deep learning for video scene detection specifically on real-world motion-picture films (as opposed to egocentric videos, sports, news, etc.).
(2) We present three configurations for combining learning into the OSG pipeline, with varying degrees of integration and tailored losses.
(3) We evaluate the various approaches and analyze the advantages of the different techniques.

\section{Optimal Sequential Grouping}
\label{sec:osg}

In this section we detail briefly the formulation and solution of Optimal Sequential Grouping (OSG).
For more in-depth details see our previous publication \cite{ours_ism}.
Due to the generality of the approach, the description refers to a sequence of feature vectors undergoing partitioning into groups.
For the task of video scene detection each feature vector describes a shot and the groups are the resulting scenes (see details in Section \ref{sec:technical}).

\begin{figure}
\centering
\includegraphics[width=0.5\linewidth]{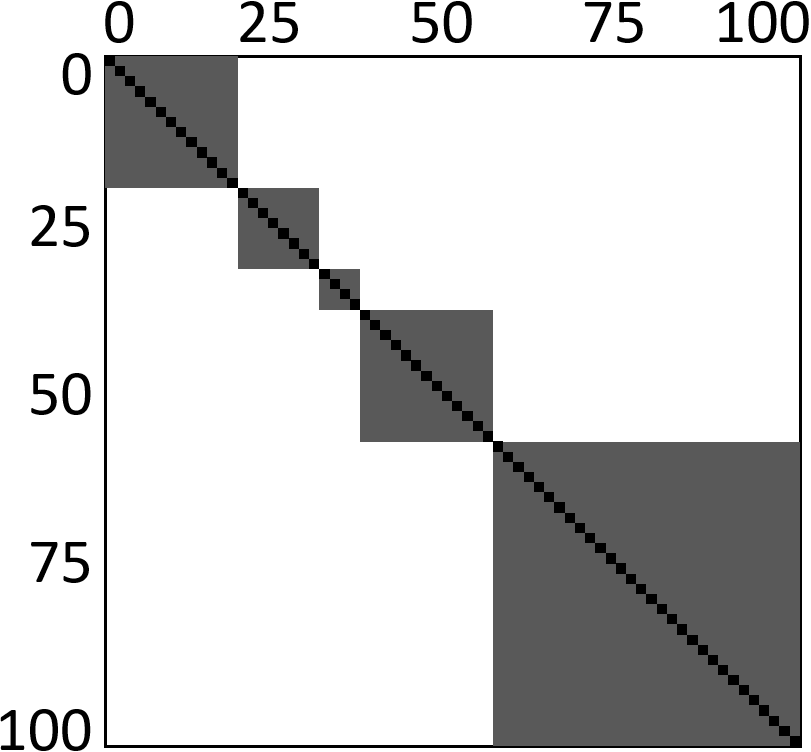}
   \caption{A depiction of how an ideal distance matrix might look for a heterogeneous video. In this depiction, higher values are assigned brighter intensity. A dark block is a sequence of shots with low intra-distances which likely indicates a scene.}
\label{fig:intuition_old}
\end{figure}

Intuitively, when representing a video containing scenes as a distance matrix, we expect to see a \textit{block-diagonal} structure (see Figure \ref{fig:intuition_old}).
This structure is formed by the fact that shots belonging to the same scene will likely have lower distance values than shots belonging to different scenes.
OSG is a dynamic programming algorithm which finds the block-diagonal with the lowest intra-scene distances.

We denote a sequence of $N$ feature vectors $X_1^N=(x_1,\ldots,x_N)$ where $x_i \in \mathbb{R}^d$, $d$ is the feature vector length.
A partitioning of the sequence into $K \leq N$ groups is given by $\bar{t}=(t_1,\ldots,t_K)$, where $t_i \in \mathbb{N}$ denotes the index of the last feature vector in group $i$.
A distance metric $\mathcal{D}(x_{j_1},x_{j_2})$ measures the dissimilarity between two feature vectors.
These distances guide a cost function $\mathcal{H}(\bar{t}) \in \mathbb{R}$ which measures the loss of a given division.
The goal of OSG is to find $t^*=\text{arg\,min}(\mathcal{H})$ as the optimal division of $X_1^N$.

The additive cost function for a given division is defined as:
\begin{equation}
    \mathcal{H}(\bar{t}) = \sum_{i=1}^K \mathop{\sum\sum}_{j_1,j_2 = t_{i-1}+1}^{t_i} \mathcal{D}(x_{j_1},x_{j_2}),
\end{equation}
where the abbreviated notation of the double sum indicates that $j_1$ and $j_2$ run from $t_{i-1}+1$ to $t_i$ each.
This cost function sums all of the intra-group distances over all of the groups in the division.
Intuitively, this cost function finds a low-valued block diagonal as illustrated in Figure \ref{fig:intuition_old}.

To find the optimal division $t^*$, we build the following recursive dynamic programming table:
\begin{equation}
\label{eq:c_table}
    \mathcal{C}(n,k) = \underset{i}{\text{min}}\left\{ \mathop{\sum\sum}_{j_1,j_2 = n}^{i} \mathcal{D}(x_{j_1},x_{j_2}) + \mathcal{C}(i+1,k-1) \right\}.
\end{equation}
Here, $\mathcal{C}(n,k)$ is the optimal cost when dividing $X_n^N$ into $k$ groups.
Essentially, we find the best cost for dividing a sub-sequence which begins at index $n$, where $i$ is the location of the first point of division for this sub-sequence.
The initialization:
\begin{equation}
    \mathcal{C}(n,1) = \mathop{\sum\sum}_{j_1,j_2 = n}^{N} \mathcal{D}(x_{j_1},x_{j_2}),
\end{equation}
is the cost of a sub-sequence starting at $n$ without any divisions.
Building the table with ascending $k=2 \ldots K$ (rising number of divisions) and descending $n=N \ldots 1$ (increasingly longer sequences) allows us to utilize the table to aggregate the partial solutions.
Therefore we have that: $\mathcal{C}(1,K)=\mathcal{H}(t^*)$, and we can reconstruct $t^*$ by storing the indexes of the chosen divisions from \eqref{eq:c_table}.

The number of divisions $K$ is estimated using the log-elbow approach \cite{ours_ism,prior_story_graphs}.
To this end, the singular values of the distance matrix are computed, and the plot of the log values is analyzed.
The point of plateau (`elbow') in the plot was shown to correspond to the number of blocks with intuition from performing a low-rank matrix approximation.
See Appendix \ref{sup:elbow} for details on how the elbow point is estimated.

When incorporating multiple modalities \cite{ours_mmsp}, the distance for each modality is used to build its own table $\mathcal{C}_x$, $\mathcal{C}_y$, where the subscript indicates the modality, and $Y_1^N=(y_1,\ldots,y_N)$ is an additional  modality.
Instead of choosing the point of division which yields the lowest cost for a single modality, the modality which has a more pronounced division point is chosen.
We define:
\begin{equation}
\label{eq:g}
    G_x^{n,k}(i) =  \mathop{\sum\sum}_{j_1,j_2 = n}^{i} \mathcal{D}_x(x_{j_1},x_{j_2}) + \mathcal{C}_x(i+1,k-1),
\end{equation}
which is the argument of the minimum function in \eqref{eq:c_table}.
$G_x^{n,k}$ is normalized to indicate the relative inclination for division:
\begin{equation}
\label{eq:g_norm}
    \hat{G}_x(i) = \frac{G_x(i)-\text{mean}\left\{G_x\right\}}{\text{std}\left\{G_x\right\}},
\end{equation}
and the index is chosen as: $\underset{i}{\text{arg\,min}}\left\{\text{min}(\hat{G}_x(i),\hat{G}_y(i))\right\}$ (superscripts were omitted for the sake of readability).

\section{Learnable OSG}
\label{sec:learnable}

Despite the strengths of OSG as an unsupervised optimization scheme, the main weakness is the dependency on choosing the representative features $X_1^N$ and distance metric $\mathcal{D}$.
Here, deep learning as a data representation mechanism can be a powerful tool when joined with OSG.
In this section we detail three possible configurations for joining learning with the OSG algorithm.
In all of the sections below, we take the shot representations $X_1^N$ and feed them through a series of fully connected layers to learn a new representation $\widetilde{X}_1^N$ (in the notations below, we omit the tilde for simplicity).
These parameters are what the network learns to better perform OSG.

\subsection{Cluster Embedding (OSG-Triplet)}
\label{sec:triplet}

The most direct way to apply learning to the OSG problem would be with learning an embedding.
Specifically, the triplet loss \cite{facenet} learns a feature space embedding where samples from the same class are close in the feature space while samples from different classes are further apart.
This is useful for a range of tasks, but for scene division this is doubly intuitive because the triplet loss causes samples (shots, in this case) to cluster together (see Appendix \ref{sup:triplet}).

These clusters will likely make scene detection a much simpler task, because often the task is approached as a variant of a shot clustering problem.
In an embedding where shots are clustered into scenes, we can assume that the distance matrix will possess beneficial properties for OSG.
Likely, the intra-scene distances will be reduced compared to the inter-scene distances causing the dynamic programming algorithm to arrive at the correct divisions.

Given a label $L(x_i) \in [1,K]$ indicating the number of the scene that feature vector $x_i$ belongs to, the neural network parameters are learned by minimizing the triplet loss:
\begin{equation}
   \sum \text{min} (\mathcal{D}(x_i,x_i^p)-\mathcal{D}(x_i,x_i^n)+\alpha,0).
\end{equation}
For anchor samples $x_i$, a positive and negative pair are chosen, where $L(x_i)=L(x_i^p)$ and $L(x_i) \neq L(x_i^n)$, and $\alpha$ is a margin parameter.
The samples are chosen using the semi-hard approach, where the triplets that satisfy the condition $\mathcal{D}(x_i,x_i^p)<\mathcal{D}(x_i,x_i^n)<\mathcal{D}(x_i,x_i^p)+\alpha$ are chosen.

As stated above, this approach is intuitive and likely to aid OSG in division.
In the next configurations we show how we go further to tailor the learning specifically for the OSG formulation.

\subsection{Block-Diagonal Loss (OSG-Block)}

In Section \ref{sec:osg}, we described the intuition behind OSG as identifying the block-diagonal structure in the distance matrix.
In this configuration, we apply a loss designated to strengthen that block-diagonal structure.

\begin{figure}
\centering
\includegraphics[width=0.5\linewidth]{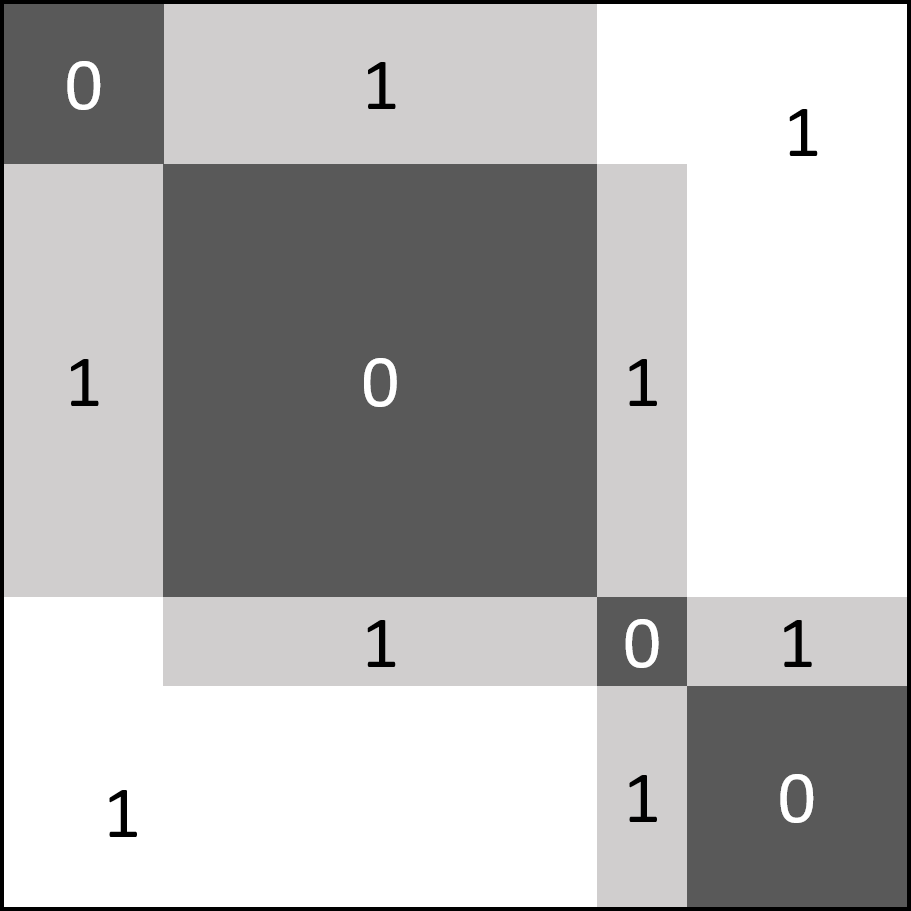}
   \caption{$D^*$. For OSG-Block the entire matrix is used, while for OSG-Block-Adjacent only the gray (dark and light) portions are considered.}
\label{fig:dlossold}
\end{figure}

If we present the distance values in a matrix $D$, where the $i$-th row and $j$-th column is $D_{i,j}=\mathcal{D}(x_i,x_j)$, then we can define an `optimal' $D^*$ as:
\begin{equation}
    D_{i,j}^*=\left\{\begin{array}{ll} 0 & L(x_i)=L(x_j) \\ 1 & \text{else} \end{array}\right. .
\end{equation}
Here, $0$ is the minimal distance and is allocated for features from the same scene, and $1$ is the maximal distance for features from different scenes (see Figure \ref{fig:dlossold}).

OSG does not need an optimal $D$ matrix to perform well.
The relative divisions are compared to each other so the correct solution only needs to have a slightly lower cost than any other solution.
However, driving $D$ toward $D^*$ will likely help OSG find the the correct division.
Therefore, the loss we use is the Frobenius norm of the subtraction:
\begin{equation}
\label{eq:dloss}
    \left\Vert D-D^* \right\Vert _F = \sqrt{\sum_i \sum_j \left| D_{i,j} - D_{i,j}^* \right|^2}.
\end{equation}

A slight variant of this loss is to not consider the inter-scene distances between scenes which are not adjacent to each other (OSG-Block-Adjacent).
The rational is that some scenes throughout a video might be quite similar to each other, but their temporal distance or an intervening scene will indicate their distinction.
The cost function in OSG accumulates the inner values of the block-diagonal, while the far off-diagonal values do not impact the decision as long as the values in between are high enough.

In this case, the loss receives only a portion of the values.
Specifically, in \eqref{eq:dloss}, we only consider values of $j$ that satisfy the constraint: $L(x_i)-1 \leq L(x_j) \leq L(x_i)+1$.
I.e., only the intra-scene distances and distances between feature vectors belonging to neighboring scenes are considered (see Figure \ref{fig:dlossold}).

\subsection{Scene Division Probabilities (OSG-Prob)}

\label{sec:osg_prob}

In this configuration, the learning process is performed through the OSG pipeline.
The OSG formulation is altered slightly to allow division probabilities to be calculated, and this is contrasted to the ground truth divisions.
The model then learns to raise the probability for division at the correct location.

In \eqref{eq:c_table}, the $C$ table is used to calculate optimal locations of division.
As in \eqref{eq:g}, we retain the relative inclinations for division.
Instead of \eqref{eq:g_norm}, we output a probability vector with the established softmin operator and aggregate the values in a larger table:
\begin{equation}
\label{eq:ctablebig}
    \mathcal{C}(n,k,i) = \frac{\text{exp}(-G^{n,k}(i))}{\sum_j \text{exp}(-G^{n,k}(j))}.
\end{equation}
The values in this table retain the probability to divide the video at point $i$ when dividing $X_n^N$ into $k$ scenes. We average these probabilities in the $C$ table over $n$ and $k$ and arrive at a vector of `scores' for division at each location in the video:
\begin{equation}
\label{eq:t}
    T(i) = \frac{1}{N \cdot K} \sum_n \sum_k  \mathcal{C}(n,k,i).
\end{equation}

Given the probabilistic nature of the values, we opt to use the cross-entropy loss on the probabilities at the indexes where a division is annotated:
\begin{equation}
    - \sum_{i \in \bar{t}_{GT}} \text{log}(T(i)).
\end{equation}
Where $\bar{t}_{GT} = \allowbreak \left\{ i | L(x_{i+1}) > L(x_i)\right\}$ is the ground truth division.

We note that there is no inherent problem with evaluating $T$ only on the ground truth division indexes.
The network cannot learn a trivial $T \equiv 1$, as high values in the $C$ table imply directly that other locations have lower values due to the softmin operation.
For a location to arrive at a high average probability, it means there must be a comprehensive indication for a division at that index, and this is what the configuration attempts to create by learning.

\section{Evaluation and Analysis}

In this section we evaluate and analyze the performance of the proposed configurations.

\subsection{Technical Details}
\label{sec:technical}

We use a pre-trained Inception-v3 architecture \cite{inception} as a 2048-dimension visual backbone feature extractor from images, and we use a pre-trained VGGish network \cite{vggish} to encode the audio segments into 128-dimension vectors.
In our experiments, we used four fully connected (FC) layers, $(3000,3000,1000,100)$ for visual and $(200,200,100,20)$ for audio.
Batch normalization was applied on all layers, and ReLU activations were applied on all layers excluding output.
The ADAM optimization algorithm was used to train the network with a learning rate of $5 \cdot 10^{-3}$.
A stopping criteria to avoid overfitting was used, and aborted the learning process when the training loss decreased to 25\% of its initial value.
The cosine distance normalized between 0 and 1 was used as $\mathcal{D}$, the margin $\alpha$ was chosen as $0.5$, and the log-elbow approach was used to estimate the number of scenes $K$.

Regarding runtime constraints, complexity of the OSG stage is unchanged compared to the original publication \cite{ours_ism}. The embedding network is relatively light-weight and the addition to the forward pass compared to the backbone is negligible. Training took roughly 24 hours on a single GPU.

For video scene detection we used the OVSD dataset \cite{ours_mmsp}.
The dataset contains 21 full-length motion-picture films from a variety of genres with ground truth scene labeling (see Appendix \ref{sup:dataset} for dataset details).
For each video, we perform shot boundary detection \cite{sbd} and extract a center image for the visual representation fed into the Inception network.
Audio for each 0.96 seconds was encoded using the VGGish network and average pooling was applied to encode each shot with its relevant audio representation.
These features were used to provide a fair comparison to \cite{ours_icmr}.
Better performance can likely be attained by incorporating advanced representations such as an I3D network \cite{i3d}.

In order to compare results on the entire OVSD dataset, we aimed to show `test' performance on all of the videos.
To accomplish this, we incorporated a 5-fold testing approach, where the videos were split into five groups of roughly equal size with regard to number of shots, i.e., some groups consisted of fewer but longer videos while others consisted of more videos, each with less shots (see Table \ref{tab:results} in the video name subscripts for the division to groups).
Five separate models were each trained on four-fifths of the data, leaving out one group for testing.
At test time, the five models were applied each to its test group, and scores were averaged over all of the videos.

\subsection{Configuration Analysis}

In this section, we analyze the behavior of the various configurations.

\begin{table*}[t]
    \caption{An example $D$ from OVSD. On the left: Ground Truth ($D^*$), Orig (without an applied embedding), Epoch 0 (embedding before learning). On the right, trained examples after 20 epochs for: OSG-Triplet, OSG-Block, OSG-Block-Adjacent, and OSG-Prob, with corresponding gradients (bottom row)}
    \centering
    \begin{tabular}{cc|}
        \multicolumn{2}{c|}{Ground}  \\
        \multicolumn{2}{c|}{Truth}  \\
        \multicolumn{2}{c|}{\raisebox{-.5\height}{\includegraphics[width=0.13\linewidth]{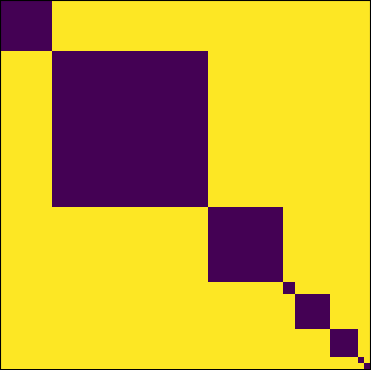}}} \\
        Orig &  Epoch 0 \\
        \raisebox{-.5\height}{\includegraphics[width=0.13\linewidth]{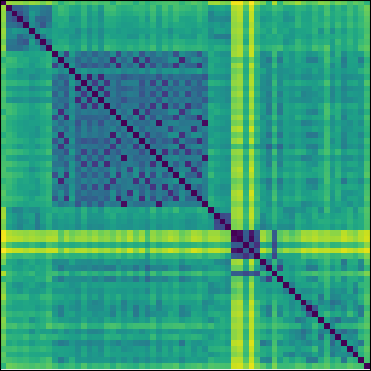}} & \raisebox{-.5\height}{\includegraphics[width=0.13\linewidth]{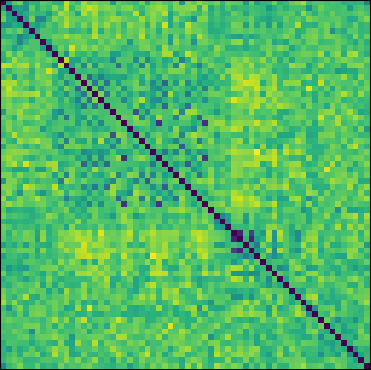}}
    \end{tabular}
    \begin{tabular}{|ccccc}
         & & OSG-Block- & \\
        OSG-Triplet & OSG-Block & Adjacent & OSG-Prob \\
        \raisebox{-.5\height}{\includegraphics[width=0.13\linewidth]{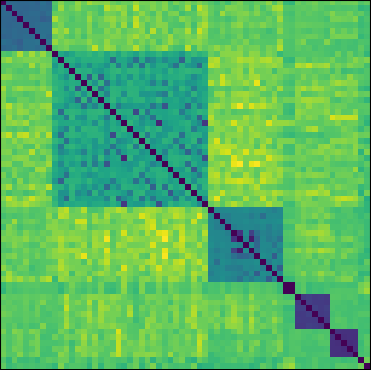}} & \raisebox{-.5\height}{\includegraphics[width=0.13\linewidth]{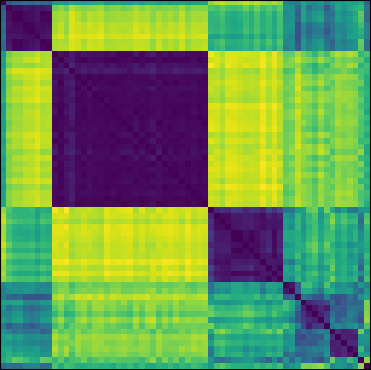}} & \raisebox{-.5\height}{\includegraphics[width=0.13\linewidth]{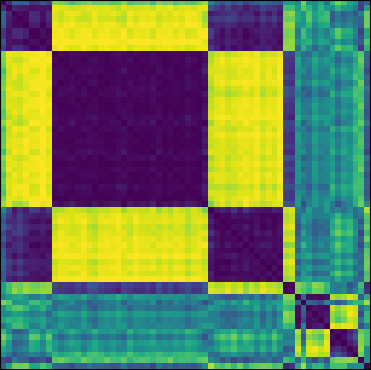}} &
        \raisebox{-.5\height}{\includegraphics[width=0.13\linewidth]{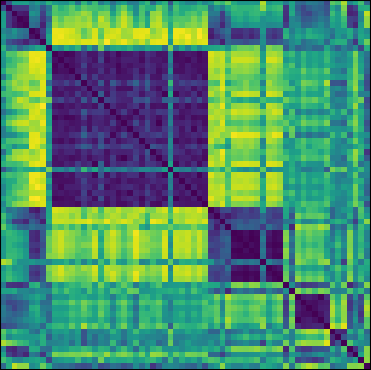}} \\
        Grad & Grad & Grad & Grad \\ \raisebox{-.5\height}{\includegraphics[width=0.13\linewidth]{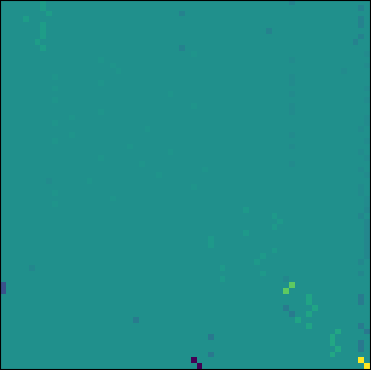}} & \raisebox{-.5\height}{\includegraphics[width=0.13\linewidth]{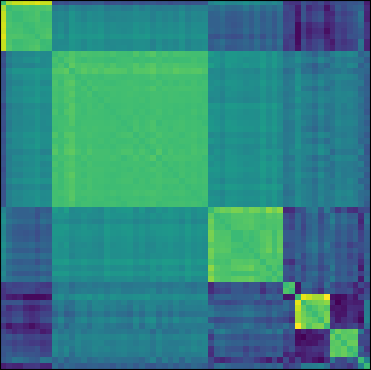}} & \raisebox{-.5\height}{\includegraphics[width=0.13\linewidth]{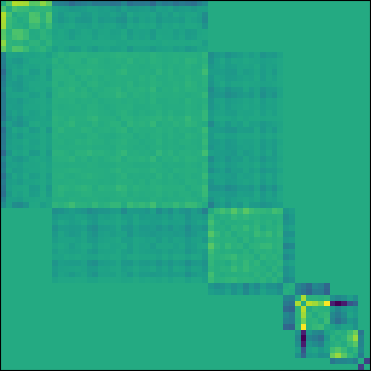}} &
        \raisebox{-.5\height}{\includegraphics[width=0.13\linewidth]{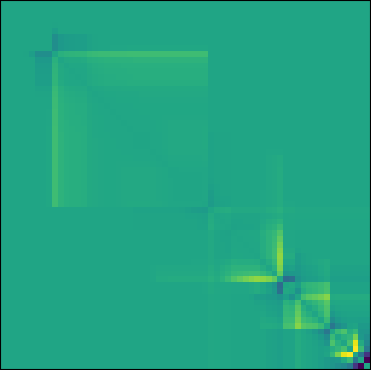}}
    \end{tabular}
    \label{tab:dexamples}
\end{table*}

In Table \ref{tab:dexamples}, we present various stages of $D$ from visual features of the video Meridian from OVSD with the accompanied ground truth $D^*$.
This video contains the smallest amount of shots, and offers the ability to visually and qualitatively inspect the structure of the matrix $D$ and behavior of the configurations.
`Orig' displays the distance metric applied directly to the backbone features.
Since the features were chosen to provide a fair comparison, this is exactly the matrix that the method in \cite{ours_icmr} applies OSG on.
`Epoch 0' is the matrix from the features after applying an untrained embedding network and incurring a level of noise.
On the right are the matrix after 20 epochs under the different configurations, with the gradients below each matrix.

OSG-Triplet and OSG-Block both explicitly strive to minimize the distances between shots from the same scene and raise the distances between shots belonging to different scenes.
The main difference between these approaches is strengthening the block-diagonal structure to better help OSG performance.
While the triplet loss focuses on distinct samples, the block-diagonal loss concentrates on the complete structure and raises the values outside of the block-diagonal.

It is interesting to note the balance between how the configurations emphasize small versus large scenes.
OSG-Triplet based off of distinct samples manages to emphasize the small scenes and has difficulty with long scenes.
OSG-Block-Adjacent compared to OSG-Block dismisses the far off-diagonal blocks, focuses on the areas which are more important (the areas between adjacent scenes), and manages to accentuate both large and small scenes.
OSG-Prob seems to converge more slowly but consistently over the scenes. 

One interesting aspect to explore is how the losses affect the gradients of the distance matrix.
In Table \ref{tab:dexamples} (right, bottom), we see the map of the gradient values for the various configurations.
As can be expected, OSG-Triplet is dependant on individual values of distance, OSG-Block puts emphasis on the entire block-diagonal, and OSG-Block-Adjacent on the relevant section of the block-diagonal as defined by the ground truth,  while OSG-Prob has a much more `local' impact focused around the points of division.
For OSG-Prob, this is a direct outcome of the formulation which emphasizes the value of the average probability on the scene division.

OVSD \cite{ours_mmsp}, is one of the only freely-available video scene detection datasets.
Despite the substantial length and variety, the amount of data is still very limited especially when considering other deep learning tasks.
The reason the network is able to learn at all, we assume, is because the configurations we chose do not treat a complete video as a sample, but rather a shot as a sample.
With each scene acting as a label instead of the entire division being considered a label, the models manage to generalize the important elements which represent shots belonging to the same scene.

\begin{figure*}[t]
\centering
\includegraphics[width=\linewidth]{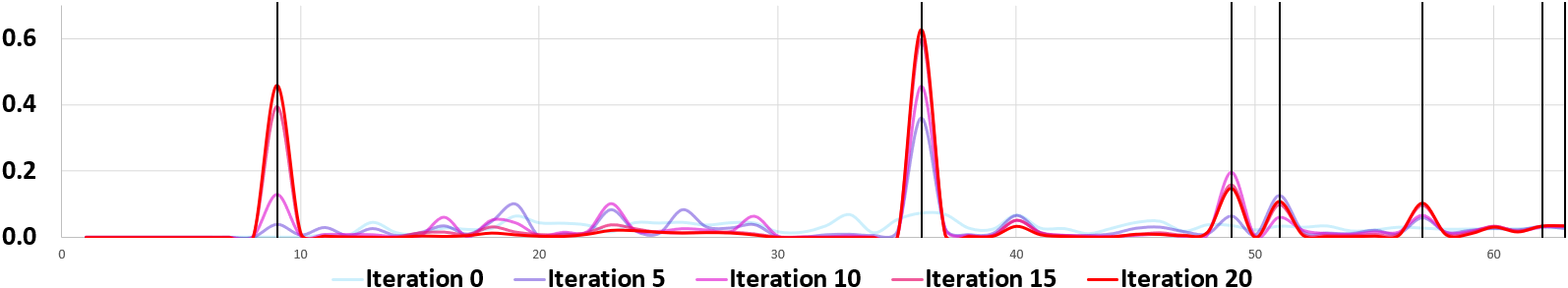}
   \caption{Progression of $T$ as a function of $i$ (shot number) over a number of iterations. Graphs go from translucent blue to opaque red as iterations progress (best viewed in color). Vertical black lines indicate ground truth divisions.}
\label{fig:t_evo}
\end{figure*}

Regarding the behavior of $T$, in Figure \ref{fig:t_evo} we show the progression of the values of $T(i)$ over a number of iterations.
We can see that as the iterations progress, $T$ raises the probability at the ground truth points of division.
The probability is lowered for locations with no true division even though this is not specifically enforced by the loss but rather an outcome of the construction of $T$ (see Section \ref{sec:osg_prob}).
Additionally we see that the model has difficulty on the last small scenes which are more difficult to enforce.

\subsection{Baselines}

\label{sec:baselines}

As explained in Section \ref{sec:intro}, the motivation to integrate learning into the OSG pipeline is to leverage the strengths of the deterministic formulations together with the ability to learn from examples.
To emphasise this point, besides comparing to state-of-the-art methods, we implement two `pure' deep learning baselines.

The first is a sliding window approach. Using a window of $W$ consecutive feature vectors (representing shots), the network is trained to identify when the scene transition is precisely in the middle of the window.
This is a naive but straightforward way to apply learning to the problem but without leveraging a formulation such as OSG.
To represent a fair comparison, we used all the same parameters as the embedding network detailed above.
$W=4$ embedded feature vectors were concatenated and a final FC layer to size $1$ was added followed by a sigmoid output (additional values of $W$ were tested resulting in comparable or worse performance).

The second baseline we used is a Long Short-Term Memory (LSTM) recurrent neural network architecture.
The LSTM is a classic choice for modeling sequence-to-sequence problems, and presents a more advanced baseline for comparison.
A bi-directional LSTM component was used with a hidden state of length $1000$, followed by two FC layers sized $100$ and $1$, the former with a ReLU activation and the latter with a sigmoid output.
The network processes the sequence of feature vectors and for each time step outputs the probability for the current feature representing the end of a scene.

Both baselines were trained using the same methodology as our configurations.
As an unfair advantage, sigmoid threshold was chosen as the value which maximized test performance.
Table \ref{tab:results} under the `Supervised' column shows the sliding window (SW) and LSTM results.

\subsection{Scene Detection Evaluation}

\begin{figure*}[t]
\centering
\includegraphics[width=\linewidth]{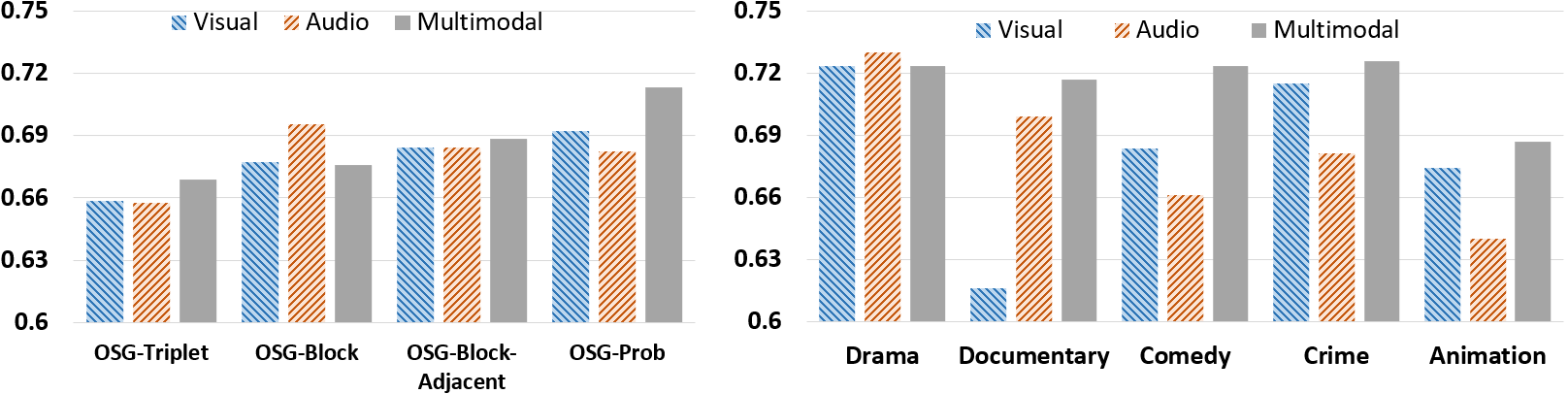}
   \caption{Modality analysis. Average $F$-score when using visual, audio, or a multimodal fusion. Performance per configuration (left), and performance of OSG-Prob consolidated per genre (right).}
\label{fig:barchart}
\end{figure*}

We measure the performance of our OSG configurations on the OVSD dataset \cite{ours_mmsp}.
For a metric, we use the widely accepted Coverage $C$ and Overflow $O$ \cite{coverageoverflow}, with a single value $F$-score for assessing the quality of division as the harmonic mean between $C$ and $1-O$.

Figure \ref{fig:barchart} on the left presents the average $F$-score for the various configurations on the tested videos.
We present the performance when using the visual or audio features separately, and when performing OSG with the multimodal fusion approach (see Section \ref{sec:osg}).
The results show superior performance for OSG-Prob, and specifically the multimodal fusion approach, which is more preferable than using a single modality for most of the configurations.

As an analysis of the behavior of each modality, we divided the performance of OSG-Prob per genre of OVSD (Figure \ref{fig:barchart} on the right).
It can be noted that for documentaries where it is characteristic for the visuals to change often, but for speakers to stay constant within a scene, the audio modality played a vital role in division.
On the other hand, for comedy, crime, and animation, the visual aspect played a slightly more important role.
In drama, both modalities contributed greatly, which coincides with the fact that often the changing visuals are accompanied with matching auditory ambiance in this genre.

\begin{table*}[t]
\begin{center}
\caption{$F$-score results on OVSD. Best score per video in bold. Subscript on video name indicates the group for 5-fold testing}
\label{tab:results}
\begin{tabular}{c|cccc|cc|ccc|cc}
& \multicolumn{4}{c|}{This Work} & \multicolumn{2}{c|}{Supervised} & \multicolumn{3}{c|}{OSG Prior Art} & \multicolumn{2}{c}{Prior Art} \\
\hline
& OSG- & OSG- & OSG- & OSG- & & & & & & & \\
Video Name		&Prob	&Block-	&Block	&Triplet & SW & LSTM &\cite{ours_icmr}		& \cite{ours_mmsp} &\cite{ours_ism} &\cite{mklab}	&\cite{lb} \\
& & Adjacent & & & & & & & & & \\
\hline
1000$_1$		&0.70			&\textbf{0.74}	&0.67			&0.68			&0.49	&0.38	&0.60	&0.38			&0.57			&0.50	&0.39			\\
BBB$_5$			&0.76			&0.74			&0.77			&0.81			&0.66	&0.54	&0.63	&\textbf{0.83}	&0.69			&0.49	&0.46   		\\
BWNS$_1$		&\textbf{0.80}	&0.74			&0.71			&0.75			&0.65	&0.62	&0.70	&0.63			&0.20			&0.61	&0.43   		\\
CH7$_1$			&0.72			&\textbf{0.74}	&0.66			&0.67			&0.56	&0.45	&0.60	&0.63			&0.49			&0.52	&0.26   		\\
CL$_2$			&\textbf{0.68}	&0.55			&0.57			&0.49			&0.60	&0.43	&0.51	&0.53			&0.53			&0.45	&0.07   		\\
ED$_2$			&0.70			&0.68			&0.64			&\textbf{0.73}	&0.55	&0.31	&0.61	&0.6			&0.69			&0.56	&0.55   		\\
FBW$_5$			&0.76			&0.77			&\textbf{0.78}	&0.76			&0.59	&0.63	&0.59	&0.57			&0.14			&0.61	&0.52   		\\
Honey$_2$		&\textbf{0.74}	&0.66			&0.67			&0.73			&0.58	&0.26	&0.63	&0.58			&0.38			&0.38	&0.36   		\\
JW$_2$			&\textbf{0.79}	&0.65			&0.65			&0.65			&0.62	&0.29	&0.63	&0.75			&0.64			&0.28	&0.22   		\\
LCDP$_5$		&\textbf{0.73}	&0.60			&0.60			&0.61			&0.61	&0.45	&0.72	&0.53			&0.42			&0.18	&0.22   		\\
LM$_5$			&0.73			&\textbf{0.74}	&0.64			&0.65			&0.60	&0.60	&0.64	&0.69			&0.25			&0.71	&0.28   		\\
Meridian$_2$	&0.64			&0.69			&0.68			&0.69			&0.47	&0.66	&0.79	&0.63			&0.71			&0.63	&\textbf{0.82}	\\
Oceania$_3$		&0.73			&0.77			&\textbf{0.78}	&0.68			&0.40	&0.62	&0.61	&0.67			&0.45			&0.51	&0.26			\\
Pentagon$_3$	&0.65			&0.65			&0.64			&\textbf{0.80}	&0.57	&0.61	&0.65	&0.73			&0.18			&0.48	&0.16			\\
Route 66$_3$	&0.67			&0.71			&0.66			&\textbf{0.72}	&0.44	&0.19	&0.60	&0.54			&0.31			&0.36	&0.05			\\
SDM$_4$			&\textbf{0.82}	&0.72			&0.80			&\textbf{0.82}	&0.51	&0.53	&0.70	&0.68			&0.67			&0.81	&0.55			\\
Sintel$_5$		&0.59			&0.59			&0.60			&0.48			&0.43	&0.54	&0.58	&0.46			&\textbf{0.66}	&0.59	&0.51			\\
SStB$_3$		&\textbf{0.66}	&0.51			&0.65			&0.44			&0.34	&0.51	&0.62	&0.46			&0.48			&0.43	&0.22			\\
SW$_4$			&\textbf{0.72}	&0.71			&0.67			&0.66			&0.52	&0.59	&0.61	&0.55			&0.36			&0.40	&0.13			\\
ToS$_5$			&0.66			&\textbf{0.78}	&0.56			&0.53			&0.65	&0.55	&0.73	&0.5			&0.62			&0.75	&0.23			\\
Valkaama$_5$	&0.74			&0.71			&\textbf{0.78}	&0.69			&0.66	&0.46	&0.70	&0.63			&0.73			&0.73	&0.16			\\
\hline
\hline
Average		&\textbf{0.71}	&0.69	&0.68	&0.67	&0.55	&0.49	&0.64	&0.60	&0.52	&0.46	&0.33	\\
\hline
\end{tabular}
\end{center}
\end{table*}

Table \ref{tab:results} presents the $F$-scores over all of the videos in the OVSD dataset with our OSG configurations leveraging multimodal fusion of visual and audio features.
As a comparison, we show the `Prior Art' column which are two state-of-the-art unsupervised methods \cite{lb,mklab}.
The `OSG Prior Art' column \cite{ours_icmr,ours_mmsp,ours_ism}, are the prior art on OSG.
Specifically, as stated above in Section \ref{sec:technical}, our backbone features are the same as \cite{ours_icmr}.
Therefore, this can be seen as directly comparable to the case when no learning is performed.
Additionally, the `Supervised' column presents the results of our implemented baselines as detailed in Section \ref{sec:baselines}.
These represent applying learning to the problem without leveraging a strong deterministic formulation such as OSG.

\begin{figure*}[t]
\centering
\includegraphics[width=0.97\linewidth]{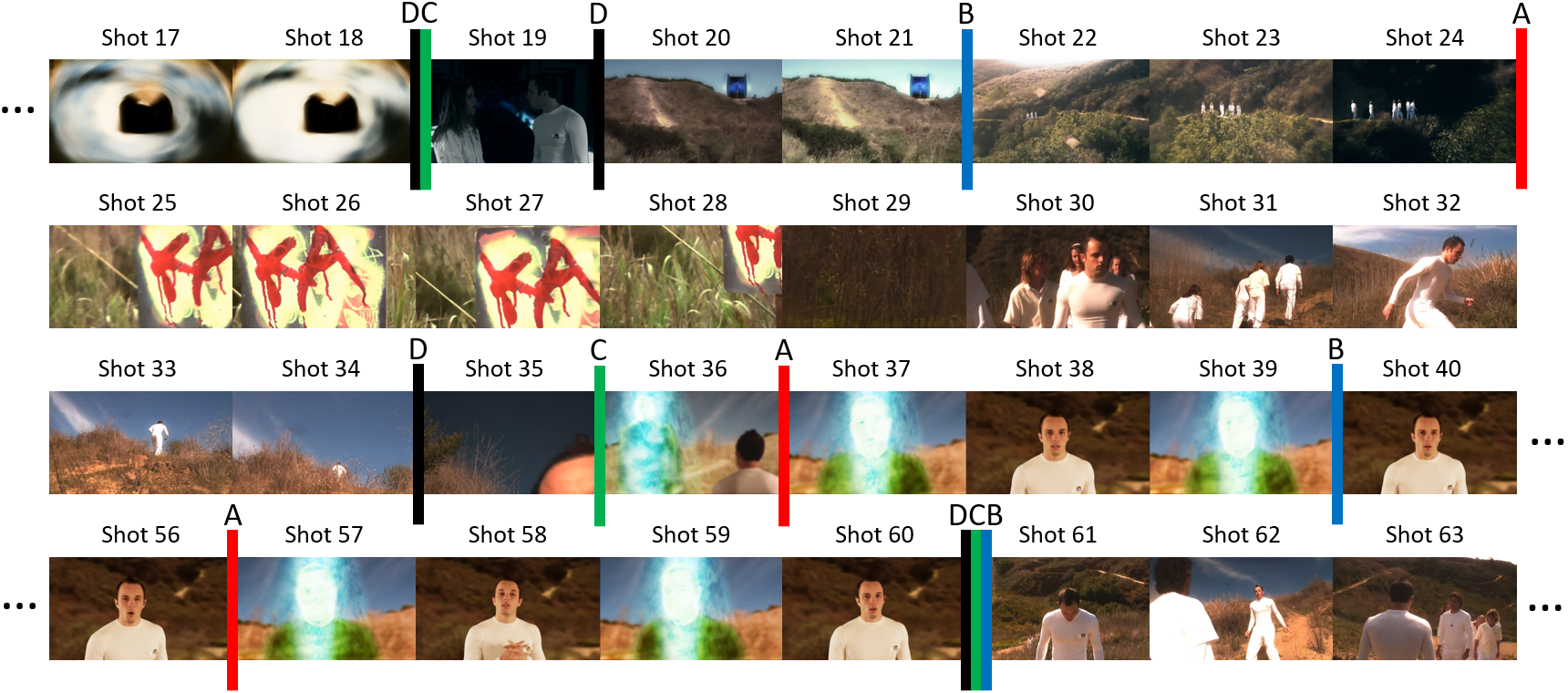}
   \caption{Qualitative results on sections of Jathia's Wager from OVSD dataset. Division marked by A OSG-Triplet (red) B OSG-Block (blue) C OSG-Prob (green) and D Ground truth (black) - See Appendix \ref{sup:visual}. Images © 2009, Solomon Rothman.}
\label{fig:frames}
\end{figure*}

It is interesting to note the substantial increase in performance of the configurations compared to the prior art on OSG.
But even more so, when comparing to the performance of the supervised baselines, we can clearly see the benefit of leveraging both learning and OSG.
Despite OSG-Prob attaining the best results, it is not directly our intention to promote it as the only viable option.
Indeed, with closely comparable results, we felt that it would be beneficial for the advancement of future work to present multiple options, as opposed to promoting a single architecture.
When extending or applying our work to future problems, it can be beneficial to choose the relevant configuration
for the problem while being able to understand the behavior and trade-offs.

Figure \ref{fig:frames} shows results on part of a video from the OVSD dataset (additional results in Appendix \ref{sup:visual}).
In general, we can see divisions which result in reasonable and often precise scene divisions.
Using these divisions for applying video understanding and classification technologies will undoubtedly be superior over applying them on the entire video or on naive uniform divisions.

\section{Conclusion}

In this work, we have presented a novel approach for incorporating learning into OSG for the task of video scene detection.
We presented a number of different configurations, evaluated their performance, and analyzed their behavior and various advantages.
Overall, our goal was to explore the possibility of integrating learning into the powerful formulation of OSG, so as to merge learning models with this effective analytic algorithm.
We demonstrated this ability with varying amounts of model complexity and dependence on the OSG pipeline.
Beyond creating a more precise and robust video scene detection technology, we believe this approach can enable constructing a model with designated performance on specific content or genres.
Additionally, this model could be integrated into other video models or leveraged for additional video understanding tasks.
We hope this work encourages continued research in the field of temporal video analysis.

\bibliographystyle{ACM-Reference-Format}
\bibliography{my_bib}


\begin{thebibliography}{39}


\ifx \showCODEN    \undefined \def \showCODEN     #1{\unskip}     \fi
\ifx \showDOI      \undefined \def \showDOI       #1{#1}\fi
\ifx \showISBNx    \undefined \def \showISBNx     #1{\unskip}     \fi
\ifx \showISBNxiii \undefined \def \showISBNxiii  #1{\unskip}     \fi
\ifx \showISSN     \undefined \def \showISSN      #1{\unskip}     \fi
\ifx \showLCCN     \undefined \def \showLCCN      #1{\unskip}     \fi
\ifx \shownote     \undefined \def \shownote      #1{#1}          \fi
\ifx \showarticletitle \undefined \def \showarticletitle #1{#1}   \fi
\ifx \showURL      \undefined \def \showURL       {\relax}        \fi
\providecommand\bibfield[2]{#2}
\providecommand\bibinfo[2]{#2}
\providecommand\natexlab[1]{#1}
\providecommand\showeprint[2][]{arXiv:#2}

\bibitem[\protect\citeauthoryear{Apostolidis and Mezaris}{Apostolidis and
  Mezaris}{2014}]%
        {sbd2}
\bibfield{author}{\bibinfo{person}{Evlampios Apostolidis} {and}
  \bibinfo{person}{Vasileios Mezaris}.} \bibinfo{year}{2014}\natexlab{}.
\newblock \showarticletitle{Fast shot segmentation combining global and local
  visual descriptors}. In \bibinfo{booktitle}{\emph{2014 IEEE International
  Conference on Acoustics, Speech and Signal Processing (ICASSP)}}. IEEE,
  \bibinfo{pages}{6583--6587}.
\newblock


\bibitem[\protect\citeauthoryear{Baraldi, Grana, and Cucchiara}{Baraldi
  et~al\mbox{.}}{2015a}]%
        {lb}
\bibfield{author}{\bibinfo{person}{Lorenzo Baraldi},
  \bibinfo{person}{Costantino Grana}, {and} \bibinfo{person}{Rita Cucchiara}.}
  \bibinfo{year}{2015}\natexlab{a}.
\newblock \showarticletitle{Analysis and Re-Use of Videos in Educational
  Digital Libraries with Automatic Scene Detection}. In
  \bibinfo{booktitle}{\emph{11th Italian Research Conference on Digital
  Libraries}}. \bibinfo{publisher}{Springer}, \bibinfo{pages}{155--164}.
\newblock


\bibitem[\protect\citeauthoryear{Baraldi, Grana, and Cucchiara}{Baraldi
  et~al\mbox{.}}{2015b}]%
        {prior_lb_siam}
\bibfield{author}{\bibinfo{person}{Lorenzo Baraldi},
  \bibinfo{person}{Costantino Grana}, {and} \bibinfo{person}{Rita Cucchiara}.}
  \bibinfo{year}{2015}\natexlab{b}.
\newblock \showarticletitle{A Deep Siamese Network for Scene Detection in
  Broadcast Videos}. In \bibinfo{booktitle}{\emph{Proceedings of the 23rd ACM
  International Conference on Multimedia}} (Brisbane, Australia)
  \emph{(\bibinfo{series}{MM '15})}. \bibinfo{publisher}{ACM},
  \bibinfo{address}{New York, NY, USA}, \bibinfo{pages}{1199--1202}.
\newblock
\showISBNx{978-1-4503-3459-4}
\urldef\tempurl%
\url{https://doi.org/10.1145/2733373.2806316}
\showDOI{\tempurl}


\bibitem[\protect\citeauthoryear{Baraldi, Grana, and Cucchiara}{Baraldi
  et~al\mbox{.}}{2015c}]%
        {lb_measure}
\bibfield{author}{\bibinfo{person}{Lorenzo Baraldi},
  \bibinfo{person}{Costantino Grana}, {and} \bibinfo{person}{Rita Cucchiara}.}
  \bibinfo{year}{2015}\natexlab{c}.
\newblock \showarticletitle{Measuring scene detection performance}. In
  \bibinfo{booktitle}{\emph{Iberian Conference on Pattern Recognition and Image
  Analysis}}. Springer, \bibinfo{pages}{395--403}.
\newblock


\bibitem[\protect\citeauthoryear{Baraldi, Grana, and Cucchiara}{Baraldi
  et~al\mbox{.}}{2015d}]%
        {sbd}
\bibfield{author}{\bibinfo{person}{Lorenzo Baraldi},
  \bibinfo{person}{Costantino Grana}, {and} \bibinfo{person}{Rita Cucchiara}.}
  \bibinfo{year}{2015}\natexlab{d}.
\newblock \showarticletitle{Shot and scene detection via hierarchical
  clustering for re-using broadcast video}. In
  \bibinfo{booktitle}{\emph{International Conference on Computer Analysis of
  Images and Patterns}}. Springer, \bibinfo{pages}{801--811}.
\newblock


\bibitem[\protect\citeauthoryear{Carreira and Zisserman}{Carreira and
  Zisserman}{2017}]%
        {i3d}
\bibfield{author}{\bibinfo{person}{Joao Carreira} {and} \bibinfo{person}{Andrew
  Zisserman}.} \bibinfo{year}{2017}\natexlab{}.
\newblock \showarticletitle{Quo vadis, action recognition? a new model and the
  kinetics dataset}. In \bibinfo{booktitle}{\emph{proceedings of the IEEE
  Conference on Computer Vision and Pattern Recognition}}.
  \bibinfo{pages}{6299--6308}.
\newblock


\bibitem[\protect\citeauthoryear{Chasanis, Likas, and Galatsanos}{Chasanis
  et~al\mbox{.}}{2008}]%
        {prior_sequence_alignment}
\bibfield{author}{\bibinfo{person}{Vasileios~T Chasanis},
  \bibinfo{person}{Aristidis~C Likas}, {and} \bibinfo{person}{Nikolaos~P
  Galatsanos}.} \bibinfo{year}{2008}\natexlab{}.
\newblock \showarticletitle{Scene detection in videos using shot clustering and
  sequence alignment}.
\newblock \bibinfo{journal}{\emph{IEEE transactions on multimedia}}
  \bibinfo{volume}{11}, \bibinfo{number}{1} (\bibinfo{year}{2008}),
  \bibinfo{pages}{89--100}.
\newblock


\bibitem[\protect\citeauthoryear{Del~Fabro and B{\"o}sz{\"o}rmenyi}{Del~Fabro
  and B{\"o}sz{\"o}rmenyi}{2013}]%
        {vsd_survey}
\bibfield{author}{\bibinfo{person}{Manfred Del~Fabro} {and}
  \bibinfo{person}{Laszlo B{\"o}sz{\"o}rmenyi}.}
  \bibinfo{year}{2013}\natexlab{}.
\newblock \showarticletitle{State-of-the-art and future challenges in video
  scene detection: a survey}.
\newblock \bibinfo{journal}{\emph{Multimedia systems}} \bibinfo{volume}{19},
  \bibinfo{number}{5} (\bibinfo{year}{2013}), \bibinfo{pages}{427--454}.
\newblock


\bibitem[\protect\citeauthoryear{Didona, Quaglia, Romano, and Torre}{Didona
  et~al\mbox{.}}{2015}]%
        {intro_deeplearning_analytical2}
\bibfield{author}{\bibinfo{person}{Diego Didona}, \bibinfo{person}{Francesco
  Quaglia}, \bibinfo{person}{Paolo Romano}, {and} \bibinfo{person}{Ennio
  Torre}.} \bibinfo{year}{2015}\natexlab{}.
\newblock \showarticletitle{Enhancing performance prediction robustness by
  combining analytical modeling and machine learning}. In
  \bibinfo{booktitle}{\emph{Proceedings of the 6th ACM/SPEC international
  conference on performance engineering}}. ACM, \bibinfo{pages}{145--156}.
\newblock


\bibitem[\protect\citeauthoryear{Endert, Ribarsky, Turkay, Wong, Nabney,
  Blanco, and Rossi}{Endert et~al\mbox{.}}{2017}]%
        {intro_deeplearning_analytical4}
\bibfield{author}{\bibinfo{person}{Alex Endert}, \bibinfo{person}{William
  Ribarsky}, \bibinfo{person}{Cagatay Turkay}, \bibinfo{person}{BL~William
  Wong}, \bibinfo{person}{Ian Nabney}, \bibinfo{person}{I~D{\'\i}az Blanco},
  {and} \bibinfo{person}{Fabrice Rossi}.} \bibinfo{year}{2017}\natexlab{}.
\newblock \showarticletitle{The state of the art in integrating machine
  learning into visual analytics}. In \bibinfo{booktitle}{\emph{Computer
  Graphics Forum}}, Vol.~\bibinfo{volume}{36}. Wiley Online Library,
  \bibinfo{pages}{458--486}.
\newblock


\bibitem[\protect\citeauthoryear{Furnari, Farinella, and Battiato}{Furnari
  et~al\mbox{.}}{2016}]%
        {prior_ego1}
\bibfield{author}{\bibinfo{person}{Antonino Furnari},
  \bibinfo{person}{Giovanni~Maria Farinella}, {and} \bibinfo{person}{Sebastiano
  Battiato}.} \bibinfo{year}{2016}\natexlab{}.
\newblock \showarticletitle{Temporal segmentation of egocentric videos to
  highlight personal locations of interest}. In
  \bibinfo{booktitle}{\emph{European Conference on Computer Vision}}. Springer,
  \bibinfo{pages}{474--489}.
\newblock


\bibitem[\protect\citeauthoryear{Gao, Ge, Chen, and Nevatia}{Gao
  et~al\mbox{.}}{2018}]%
        {intro_tasks_analysis}
\bibfield{author}{\bibinfo{person}{Jiyang Gao}, \bibinfo{person}{Runzhou Ge},
  \bibinfo{person}{Kan Chen}, {and} \bibinfo{person}{Ram Nevatia}.}
  \bibinfo{year}{2018}\natexlab{}.
\newblock \showarticletitle{Motion-appearance co-memory networks for video
  question answering}. In \bibinfo{booktitle}{\emph{Proceedings of the IEEE
  Conference on Computer Vision and Pattern Recognition}}.
  \bibinfo{pages}{6576--6585}.
\newblock


\bibitem[\protect\citeauthoryear{Han and Wu}{Han and Wu}{2011}]%
        {prior_dynamic_programming}
\bibfield{author}{\bibinfo{person}{Bo Han} {and} \bibinfo{person}{Weiguo Wu}.}
  \bibinfo{year}{2011}\natexlab{}.
\newblock \showarticletitle{Video scene segmentation using a novel boundary
  evaluation criterion and dynamic programming}. In
  \bibinfo{booktitle}{\emph{2011 IEEE International conference on multimedia
  and expo}}. IEEE, \bibinfo{pages}{1--6}.
\newblock


\bibitem[\protect\citeauthoryear{Haroon, Baber, Ullah, Daudpota, Bakhtyar, and
  Devi}{Haroon et~al\mbox{.}}{2018}]%
        {prior_vsd_1_bow}
\bibfield{author}{\bibinfo{person}{Muhammad Haroon}, \bibinfo{person}{Junaid
  Baber}, \bibinfo{person}{Ihsan Ullah}, \bibinfo{person}{Sher~Muhammad
  Daudpota}, \bibinfo{person}{Maheen Bakhtyar}, {and} \bibinfo{person}{Varsha
  Devi}.} \bibinfo{year}{2018}\natexlab{}.
\newblock \showarticletitle{Video Scene Detection Using Compact Bag of Visual
  Word Models}.
\newblock \bibinfo{journal}{\emph{Advances in Multimedia}}
  \bibinfo{volume}{2018} (\bibinfo{year}{2018}).
\newblock


\bibitem[\protect\citeauthoryear{Hershey, Chaudhuri, Ellis, Gemmeke, Jansen,
  Moore, Plakal, Platt, Saurous, Seybold, et~al\mbox{.}}{Hershey
  et~al\mbox{.}}{2017}]%
        {vggish}
\bibfield{author}{\bibinfo{person}{Shawn Hershey}, \bibinfo{person}{Sourish
  Chaudhuri}, \bibinfo{person}{Daniel~PW Ellis}, \bibinfo{person}{Jort~F
  Gemmeke}, \bibinfo{person}{Aren Jansen}, \bibinfo{person}{R~Channing Moore},
  \bibinfo{person}{Manoj Plakal}, \bibinfo{person}{Devin Platt},
  \bibinfo{person}{Rif~A Saurous}, \bibinfo{person}{Bryan Seybold},
  {et~al\mbox{.}}} \bibinfo{year}{2017}\natexlab{}.
\newblock \showarticletitle{CNN architectures for large-scale audio
  classification}. In \bibinfo{booktitle}{\emph{Acoustics, Speech and Signal
  Processing (ICASSP), 2017 IEEE International Conference on}}. IEEE,
  \bibinfo{pages}{131--135}.
\newblock


\bibitem[\protect\citeauthoryear{Kloss, Schaal, and Bohg}{Kloss
  et~al\mbox{.}}{2017}]%
        {intro_deeplearning_analytical1}
\bibfield{author}{\bibinfo{person}{Alina Kloss}, \bibinfo{person}{Stefan
  Schaal}, {and} \bibinfo{person}{Jeannette Bohg}.}
  \bibinfo{year}{2017}\natexlab{}.
\newblock \showarticletitle{Combining learned and analytical models for
  predicting action effects}.
\newblock \bibinfo{journal}{\emph{arXiv preprint arXiv:1710.04102}}
  (\bibinfo{year}{2017}).
\newblock


\bibitem[\protect\citeauthoryear{Liang, Zhang, Cheng, Xu, and Lu}{Liang
  et~al\mbox{.}}{2009}]%
        {prior_graph4}
\bibfield{author}{\bibinfo{person}{Chao Liang}, \bibinfo{person}{Yifan Zhang},
  \bibinfo{person}{Jian Cheng}, \bibinfo{person}{Changsheng Xu}, {and}
  \bibinfo{person}{Hanqing Lu}.} \bibinfo{year}{2009}\natexlab{}.
\newblock \showarticletitle{A novel role-based movie scene segmentation
  method}. In \bibinfo{booktitle}{\emph{Pacific-Rim Conference on Multimedia}}.
  Springer, \bibinfo{pages}{917--922}.
\newblock


\bibitem[\protect\citeauthoryear{Mahapatra, Mariappan, and Rajan}{Mahapatra
  et~al\mbox{.}}{2018}]%
        {intro_tasks_table_of_contents}
\bibfield{author}{\bibinfo{person}{Debabrata Mahapatra},
  \bibinfo{person}{Ragunathan Mariappan}, {and} \bibinfo{person}{Vaibhav
  Rajan}.} \bibinfo{year}{2018}\natexlab{}.
\newblock \showarticletitle{Automatic Hierarchical Table of Contents Generation
  for Educational Videos}. In \bibinfo{booktitle}{\emph{Companion Proceedings
  of the The Web Conference 2018}}. International World Wide Web Conferences
  Steering Committee, \bibinfo{pages}{267--274}.
\newblock


\bibitem[\protect\citeauthoryear{M{\"u}nzer and Schoeffmann}{M{\"u}nzer and
  Schoeffmann}{2018}]%
        {intro_tasks_browsing}
\bibfield{author}{\bibinfo{person}{Bernd M{\"u}nzer} {and}
  \bibinfo{person}{Klaus Schoeffmann}.} \bibinfo{year}{2018}\natexlab{}.
\newblock \showarticletitle{Video Browsing on a Circular Timeline}. In
  \bibinfo{booktitle}{\emph{International Conference on Multimedia Modeling}}.
  Springer, \bibinfo{pages}{395--399}.
\newblock


\bibitem[\protect\citeauthoryear{Ortis, Farinella, D’Amico, Addesso, Torrisi,
  and Battiato}{Ortis et~al\mbox{.}}{2017}]%
        {prior_ego3}
\bibfield{author}{\bibinfo{person}{Alessandro Ortis},
  \bibinfo{person}{Giovanni~M Farinella}, \bibinfo{person}{Valeria D’Amico},
  \bibinfo{person}{Luca Addesso}, \bibinfo{person}{Giovanni Torrisi}, {and}
  \bibinfo{person}{Sebastiano Battiato}.} \bibinfo{year}{2017}\natexlab{}.
\newblock \showarticletitle{Organizing egocentric videos of daily living
  activities}.
\newblock \bibinfo{journal}{\emph{Pattern Recognition}}  \bibinfo{volume}{72}
  (\bibinfo{year}{2017}), \bibinfo{pages}{207--218}.
\newblock


\bibitem[\protect\citeauthoryear{Panda, Kuanar, and Chowdhury}{Panda
  et~al\mbox{.}}{2017}]%
        {prior_cluster}
\bibfield{author}{\bibinfo{person}{Rameswar Panda}, \bibinfo{person}{Sanjay~K
  Kuanar}, {and} \bibinfo{person}{Ananda~S Chowdhury}.}
  \bibinfo{year}{2017}\natexlab{}.
\newblock \showarticletitle{Nystr{\"o}m Approximated Temporally Constrained
  Multisimilarity Spectral Clustering Approach for Movie Scene Detection}.
\newblock \bibinfo{journal}{\emph{IEEE Transactions on Cybernetics}}
  (\bibinfo{year}{2017}).
\newblock


\bibitem[\protect\citeauthoryear{Poleg, Arora, and Peleg}{Poleg
  et~al\mbox{.}}{2014}]%
        {prior_ego2}
\bibfield{author}{\bibinfo{person}{Yair Poleg}, \bibinfo{person}{Chetan Arora},
  {and} \bibinfo{person}{Shmuel Peleg}.} \bibinfo{year}{2014}\natexlab{}.
\newblock \showarticletitle{Temporal segmentation of egocentric videos}. In
  \bibinfo{booktitle}{\emph{Proceedings of the IEEE Conference on Computer
  Vision and Pattern Recognition}}. \bibinfo{pages}{2537--2544}.
\newblock


\bibitem[\protect\citeauthoryear{Protasov, Khan, Sozykin, and Ahmad}{Protasov
  et~al\mbox{.}}{2018}]%
        {prior_vsd_2_annotation}
\bibfield{author}{\bibinfo{person}{Stanislav Protasov},
  \bibinfo{person}{Adil~Mehmood Khan}, \bibinfo{person}{Konstantin Sozykin},
  {and} \bibinfo{person}{Muhammad Ahmad}.} \bibinfo{year}{2018}\natexlab{}.
\newblock \showarticletitle{Using deep features for video scene detection and
  annotation}.
\newblock \bibinfo{journal}{\emph{Signal, Image and Video Processing}}
  \bibinfo{volume}{12}, \bibinfo{number}{5} (\bibinfo{year}{2018}),
  \bibinfo{pages}{991--999}.
\newblock


\bibitem[\protect\citeauthoryear{Rasheed and Shah}{Rasheed and Shah}{2005}]%
        {prior_graph1}
\bibfield{author}{\bibinfo{person}{Zeeshan Rasheed} {and}
  \bibinfo{person}{Mubarak Shah}.} \bibinfo{year}{2005}\natexlab{}.
\newblock \showarticletitle{Detection and representation of scenes in videos}.
\newblock \bibinfo{journal}{\emph{IEEE transactions on Multimedia}}
  \bibinfo{volume}{7}, \bibinfo{number}{6} (\bibinfo{year}{2005}),
  \bibinfo{pages}{1097--1105}.
\newblock


\bibitem[\protect\citeauthoryear{Ray and Chakrabarti}{Ray and
  Chakrabarti}{2019}]%
        {intro_deeplearning_analytical3}
\bibfield{author}{\bibinfo{person}{Paramita Ray} {and} \bibinfo{person}{Amlan
  Chakrabarti}.} \bibinfo{year}{2019}\natexlab{}.
\newblock \showarticletitle{A Mixed approach of Deep Learning method and
  Rule-Based method to improve Aspect Level Sentiment Analysis}.
\newblock \bibinfo{journal}{\emph{Applied Computing and Informatics}}
  (\bibinfo{year}{2019}).
\newblock


\bibitem[\protect\citeauthoryear{Rotman, Porat, and Ashour}{Rotman
  et~al\mbox{.}}{2016}]%
        {ours_ism}
\bibfield{author}{\bibinfo{person}{Daniel Rotman}, \bibinfo{person}{Dror
  Porat}, {and} \bibinfo{person}{Gal Ashour}.} \bibinfo{year}{2016}\natexlab{}.
\newblock \showarticletitle{Robust and efficient video scene detection using
  optimal sequential grouping}. In \bibinfo{booktitle}{\emph{2016 IEEE
  International Symposium on Multimedia (ISM)}}. IEEE,
  \bibinfo{pages}{275--280}.
\newblock


\bibitem[\protect\citeauthoryear{Rotman, Porat, and Ashour}{Rotman
  et~al\mbox{.}}{2017}]%
        {ours_mmsp}
\bibfield{author}{\bibinfo{person}{Daniel Rotman}, \bibinfo{person}{Dror
  Porat}, {and} \bibinfo{person}{Gal Ashour}.} \bibinfo{year}{2017}\natexlab{}.
\newblock \showarticletitle{Robust video scene detection using multimodal
  fusion of optimally grouped features}. In \bibinfo{booktitle}{\emph{2017 IEEE
  19th International Workshop on Multimedia Signal Processing (MMSP)}}. IEEE,
  \bibinfo{pages}{1--6}.
\newblock


\bibitem[\protect\citeauthoryear{Rotman, Porat, Ashour, and Barzelay}{Rotman
  et~al\mbox{.}}{2018}]%
        {ours_icmr}
\bibfield{author}{\bibinfo{person}{Daniel Rotman}, \bibinfo{person}{Dror
  Porat}, \bibinfo{person}{Gal Ashour}, {and} \bibinfo{person}{Udi Barzelay}.}
  \bibinfo{year}{2018}\natexlab{}.
\newblock \showarticletitle{Optimally Grouped Deep Features Using Normalized
  Cost for Video Scene Detection}. In \bibinfo{booktitle}{\emph{Proceedings of
  the 2018 ACM on International Conference on Multimedia Retrieval}}. ACM,
  \bibinfo{pages}{187--195}.
\newblock


\bibitem[\protect\citeauthoryear{Rui, Huang, and Mehrotra}{Rui
  et~al\mbox{.}}{1999}]%
        {scene_definition}
\bibfield{author}{\bibinfo{person}{Yong Rui}, \bibinfo{person}{Thomas~S Huang},
  {and} \bibinfo{person}{Sharad Mehrotra}.} \bibinfo{year}{1999}\natexlab{}.
\newblock \showarticletitle{Constructing table-of-content for videos}.
\newblock \bibinfo{journal}{\emph{Multimedia systems}} \bibinfo{volume}{7},
  \bibinfo{number}{5} (\bibinfo{year}{1999}), \bibinfo{pages}{359--368}.
\newblock


\bibitem[\protect\citeauthoryear{Schroff, Kalenichenko, and Philbin}{Schroff
  et~al\mbox{.}}{2015}]%
        {facenet}
\bibfield{author}{\bibinfo{person}{Florian Schroff}, \bibinfo{person}{Dmitry
  Kalenichenko}, {and} \bibinfo{person}{James Philbin}.}
  \bibinfo{year}{2015}\natexlab{}.
\newblock \showarticletitle{Facenet: A unified embedding for face recognition
  and clustering}. In \bibinfo{booktitle}{\emph{Proceedings of the IEEE
  conference on computer vision and pattern recognition}}.
  \bibinfo{pages}{815--823}.
\newblock


\bibitem[\protect\citeauthoryear{Shemer, Rotman, and Shimkin}{Shemer
  et~al\mbox{.}}{2019}]%
        {intro_tasks_summarization}
\bibfield{author}{\bibinfo{person}{Yair Shemer}, \bibinfo{person}{Daniel
  Rotman}, {and} \bibinfo{person}{Nahum Shimkin}.}
  \bibinfo{year}{2019}\natexlab{}.
\newblock \showarticletitle{ILS-SUMM: Iterated Local Search for Unsupervised
  Video Summarization}.
\newblock \bibinfo{journal}{\emph{arXiv preprint arXiv:1912.03650}}
  (\bibinfo{year}{2019}).
\newblock


\bibitem[\protect\citeauthoryear{Sidiropoulos, Mezaris, Kompatsiaris, Meinedo,
  Bugalho, and Trancoso}{Sidiropoulos et~al\mbox{.}}{2011}]%
        {mklab}
\bibfield{author}{\bibinfo{person}{Panagiotis Sidiropoulos},
  \bibinfo{person}{Vasileios Mezaris}, \bibinfo{person}{Ioannis Kompatsiaris},
  \bibinfo{person}{Hugo Meinedo}, \bibinfo{person}{Miguel Bugalho}, {and}
  \bibinfo{person}{Isabel Trancoso}.} \bibinfo{year}{2011}\natexlab{}.
\newblock \showarticletitle{Temporal video segmentation to scenes using
  high-level audiovisual features}.
\newblock \bibinfo{journal}{\emph{IEEE Transactions on Circuits and Systems for
  Video Technology}} \bibinfo{volume}{21}, \bibinfo{number}{8}
  (\bibinfo{year}{2011}), \bibinfo{pages}{1163--1177}.
\newblock


\bibitem[\protect\citeauthoryear{Smeaton, Over, and Doherty}{Smeaton
  et~al\mbox{.}}{2010}]%
        {intro_sbd}
\bibfield{author}{\bibinfo{person}{Alan~F Smeaton}, \bibinfo{person}{Paul
  Over}, {and} \bibinfo{person}{Aiden~R Doherty}.}
  \bibinfo{year}{2010}\natexlab{}.
\newblock \showarticletitle{Video shot boundary detection: Seven years of
  TRECVid activity}.
\newblock \bibinfo{journal}{\emph{Computer Vision and Image Understanding}}
  \bibinfo{volume}{114}, \bibinfo{number}{4} (\bibinfo{year}{2010}),
  \bibinfo{pages}{411--418}.
\newblock


\bibitem[\protect\citeauthoryear{Szegedy, Vanhoucke, Ioffe, Shlens, and
  Wojna}{Szegedy et~al\mbox{.}}{2016}]%
        {inception}
\bibfield{author}{\bibinfo{person}{Christian Szegedy}, \bibinfo{person}{Vincent
  Vanhoucke}, \bibinfo{person}{Sergey Ioffe}, \bibinfo{person}{Jon Shlens},
  {and} \bibinfo{person}{Zbigniew Wojna}.} \bibinfo{year}{2016}\natexlab{}.
\newblock \showarticletitle{Rethinking the inception architecture for computer
  vision}. In \bibinfo{booktitle}{\emph{Proceedings of the IEEE Conference on
  Computer Vision and Pattern Recognition}}. \bibinfo{pages}{2818--2826}.
\newblock


\bibitem[\protect\citeauthoryear{Tapaswi, Bauml, and Stiefelhagen}{Tapaswi
  et~al\mbox{.}}{2014}]%
        {prior_story_graphs}
\bibfield{author}{\bibinfo{person}{Makarand Tapaswi}, \bibinfo{person}{Martin
  Bauml}, {and} \bibinfo{person}{Rainer Stiefelhagen}.}
  \bibinfo{year}{2014}\natexlab{}.
\newblock \showarticletitle{Storygraphs: visualizing character interactions as
  a timeline}. In \bibinfo{booktitle}{\emph{Proceedings of the IEEE Conference
  on Computer Vision and Pattern Recognition}}. \bibinfo{pages}{827--834}.
\newblock


\bibitem[\protect\citeauthoryear{Trojahn, Kishi, and Goularte}{Trojahn
  et~al\mbox{.}}{2018}]%
        {prior_vsd_3_lstm}
\bibfield{author}{\bibinfo{person}{Tiago~H. Trojahn},
  \bibinfo{person}{Rodrigo~M. Kishi}, {and} \bibinfo{person}{Rudinei
  Goularte}.} \bibinfo{year}{2018}\natexlab{}.
\newblock \showarticletitle{A New Multimodal Deep-learning Model to Video Scene
  Segmentation}. In \bibinfo{booktitle}{\emph{Proceedings of the 24th Brazilian
  Symposium on Multimedia and the Web}} (Salvador, BA, Brazil)
  \emph{(\bibinfo{series}{WebMedia '18})}. \bibinfo{publisher}{ACM},
  \bibinfo{address}{New York, NY, USA}, \bibinfo{pages}{205--212}.
\newblock
\showISBNx{978-1-4503-5867-5}
\urldef\tempurl%
\url{https://doi.org/10.1145/3243082.3243108}
\showDOI{\tempurl}


\bibitem[\protect\citeauthoryear{Vendrig and Worring}{Vendrig and
  Worring}{2002}]%
        {coverageoverflow}
\bibfield{author}{\bibinfo{person}{Jeroen Vendrig} {and}
  \bibinfo{person}{Marcel Worring}.} \bibinfo{year}{2002}\natexlab{}.
\newblock \showarticletitle{Systematic evaluation of logical story unit
  segmentation}.
\newblock \bibinfo{journal}{\emph{IEEE Transactions on Multimedia}}
  \bibinfo{volume}{4}, \bibinfo{number}{4} (\bibinfo{year}{2002}),
  \bibinfo{pages}{492--499}.
\newblock


\bibitem[\protect\citeauthoryear{Voulodimos, Doulamis, Doulamis, and
  Protopapadakis}{Voulodimos et~al\mbox{.}}{2018}]%
        {intro_deeplearning_computervision}
\bibfield{author}{\bibinfo{person}{Athanasios Voulodimos},
  \bibinfo{person}{Nikolaos Doulamis}, \bibinfo{person}{Anastasios Doulamis},
  {and} \bibinfo{person}{Eftychios Protopapadakis}.}
  \bibinfo{year}{2018}\natexlab{}.
\newblock \showarticletitle{Deep learning for computer vision: A brief review}.
\newblock \bibinfo{journal}{\emph{Computational intelligence and neuroscience}}
   \bibinfo{volume}{2018} (\bibinfo{year}{2018}).
\newblock


\bibitem[\protect\citeauthoryear{Yeung, Yeo, and Liu}{Yeung
  et~al\mbox{.}}{1998}]%
        {prior_stg}
\bibfield{author}{\bibinfo{person}{Minerva Yeung}, \bibinfo{person}{Boon-Lock
  Yeo}, {and} \bibinfo{person}{Bede Liu}.} \bibinfo{year}{1998}\natexlab{}.
\newblock \showarticletitle{Segmentation of video by clustering and graph
  analysis}.
\newblock \bibinfo{journal}{\emph{Computer vision and image understanding}}
  \bibinfo{volume}{71}, \bibinfo{number}{1} (\bibinfo{year}{1998}),
  \bibinfo{pages}{94--109}.
\newblock


\end{thebibliography}

\appendix

\section{Triplet Loss for Video Scene Detection}
\label{sup:triplet}

\begin{figure}
\centering
\includegraphics[width=\linewidth]{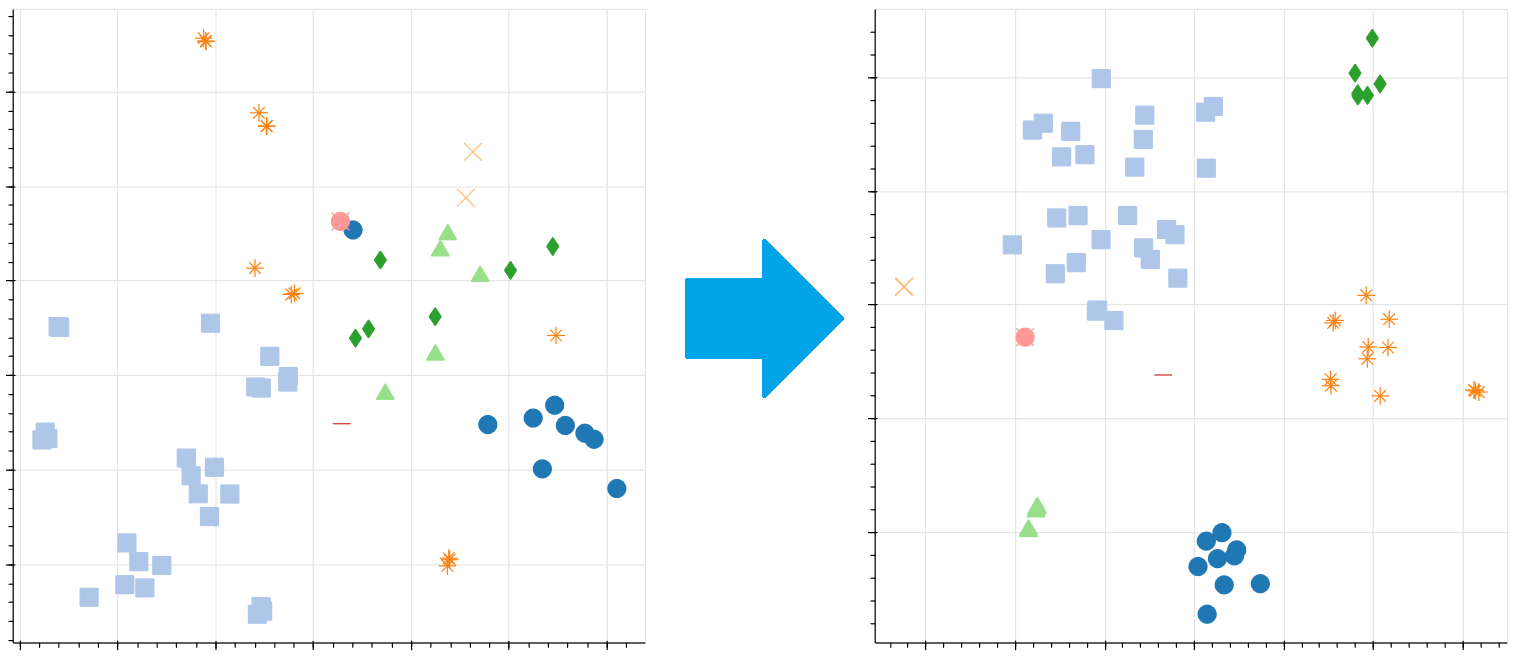}
   \caption{A reduced 2-dimensional representation of shot feature vectors (using TSNE) from the video Meridian from OVSD. Different marker types and colors represent scenes. After applying the triplet loss (right) the scenes are better separated into clusters.}
\label{fig:clusters}
\end{figure}

As mentioned in the paper, the triplet loss \cite{facenet} learns a feature space embedding where samples from the same class are close in the feature space while samples from different classes are further apart.
This is useful for a range of tasks, but for scene division this is doubly intuitive because the triplet loss causes samples (shots, in this case) to cluster together.
In Figure \ref{fig:clusters} is a reduced 2-dimensional representation of shot feature vectors (using TSNE) from the video Meridian from the OVSD dataset.
This video contains the smallest amount of shots, and offers the ability to visually and qualitatively inspect the distribution of the shot representations.

Despite the success of separating the shot representations into clusters, we can see that classic clustering algorithms might have trouble dividing correctly.
Specifically we are referring to the single-shot scenes surrounding the large light blue square scene.
In this instance, the OSG algorithm will likely be beneficial given the order and locations of the scenes, and the ability to make a decision based on the temporal order of the shots.

\section{Estimating the number of scenes $K$}
\label{sup:elbow}

The number of divisions $K$ is estimated using the log-elbow approach \cite{ours_ism,prior_story_graphs}.
To this end, the singular values of the distance matrix are computed, and the plot of the log values is analyzed.
The point of plateau (`elbow') in the plot was shown to correspond to the number of blocks with intuition from performing a low-rank matrix approximation.
The mathematical intuition is that given a distance matrix with a block-diagonal structure, we can see the rows of the matrix which belong to a specific block as being roughly linearly dependant.
If the matrix were ideal (zeros on the block diagonal and ones outside), the rank of the matrix would be exactly the number of blocks in the block diagonal.
Given a real noisy matrix, we expect the noise to act as the high frequency and low energy additions to the underlying inherent structure of the matrix.
By identifying the plateau point of the singular values we can estimate the rank of the fundamental structure of the matrix.

\begin{figure}
\centering
\includegraphics[width=0.8\linewidth]{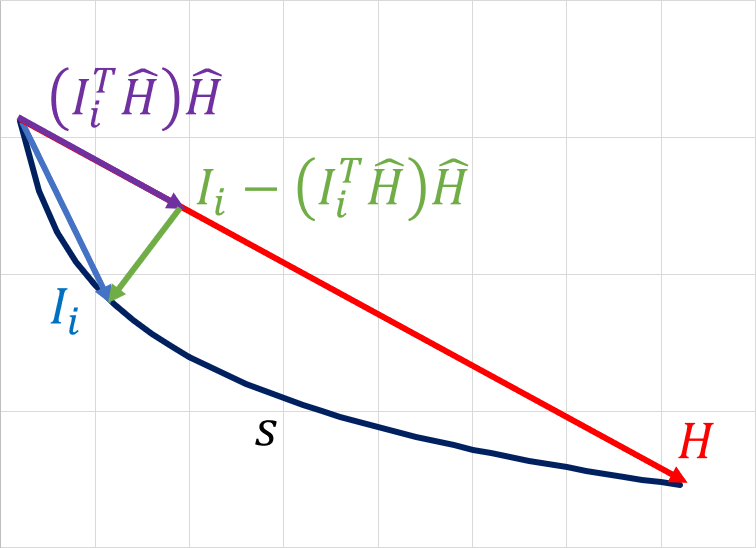}
   \caption{A depiction of the estimation of the log-elbow plateau point in the log graph of the singular values of the distance matrix.}
\label{fig:log_elbow}
\end{figure}

Practically, this plateau point is located with an elbow estimation, as the point farthest from the diagonal running over the graph.
Formally, if $s$ is the log singular values of length $N$ and we consider the index of each value as the first dimension, then the vector $I_i=[i,s_i]^T$ represents the values of the graph.
The diagonal would be: $H=[N-1,s_N-s_1]^T$, with $\hat{H} = H / \Vert H \Vert$ the unit vector in the same direction, and using the euclidean distance to each point and projecting the vector $I$, we can identify the index of the plateau point:
\begin{equation}
    \text{log-elbow}=\underset{i}{\text{arg\,max}}\left\{\Vert I_i- (I_i^T \hat{H}) \hat{H} \Vert \right\}.
\end{equation}
See Figure \ref{fig:log_elbow} for an illustration.

\section{OVSD Dataset}
\label{sup:dataset}

For video scene detection we used the OVSD dataset \cite{ours_mmsp}.
OVSD, is one of the only freely-available video scene detection datasets allowing both academic and industrial research use (creative commons licenses).
To the extent of our knowledge, this dataset is the only video scene detection dataset that has entire movies and is freely available with only minimal legal restrictions.

The dataset contains 21 full-length motion-picture films from a variety of genres with ground truth scene labeling.
Table \ref{tab:ovsd} presents the details of the OVSD dataset.
`Short Name' is the name presented in the results table in the paper to conserve space, and `\# shots' is the number of shots as estimated using a shot boundary detection method \cite{sbd}.
Some videos are defined by a number of genres (as is acceptable with films).
For the analysis per genre in the paper, the first genre was used to aggregate results, where Meridian was added to Crime (being the genre closest to Mystery).

\begin{table*}[hb]
    \centering
     \caption{OVSD dataset details}
    \begin{tabular}{c|ccccc}
        \hline
        Video & Short & Duration & \# & \# &   \\
        Name & Name & (minutes) & Scenes & Shots & Genre \\
        \hline
        \hline
        1000 Days & 1000 & 43 & 23 & 404 & Drama \\
        Big Buck Bunny & BBB & 8 & 13 & 129 & Animation \\
        Boy Who Never Slept & BWNS & 69 & 23 & 336 & Comedy, Romance \\
        CH7 & CH7 & 86 & 45 & 1293 & Crime \\
        Cosmos Laundromat & CL & 10 & 6 & 94 & Animation \\
        Elephants Dream & ED & 9 & 8 & 128 & Animation \\
        Fires Beneath Water & FBW & 76 & 63 & 411 & Documentary \\
        Honey & Honey & 86 & 21 & 326 & Drama \\
        Jathia's Wager & JW & 21 & 16 & 177 & Drama, Sci-Fi \\
        La Chute D'une Plume & LCDP & 10 & 11 & 88 & Animation \\
        Lord Meia & LM & 37 & 28 & 333 & Crime, Comedy \\
        Meridian & Meridian & 12 & 10 & 64 & Mystery, Sci-Fi \\
        Oceania & Oceania & 54 & 32 & 253 & Drama, Mystery \\
        Pentagon & Pentagon & 50 & 32 & 305 & Comedy, Drama \\
        Route 66 & Route 66 & 103 & 56 & 1357 & Documentary \\
        Seven Dead Men & SDM & 57 & 35 & 167 & Crime \\
        Sintel & Sintel & 12 & 7 & 198 & Animation \\
        Sita Sings the Blues & SStB & 81 & 53 & 1384 & Animation, Comedy \\
        Star Wreck & SW & 103 & 56 & 1439 & Comedy, Sci-Fi \\
        Tears of Steal & ToS & 10 & 6 & 136 & Drama, Sci-Fi \\
        Valkaama & Valkaama & 93 & 49 & 714 & Drama \\
        \hline
    \end{tabular}
    \label{tab:ovsd}
\end{table*}

\section{Evaluation Metric}

We measure the performance of our OSG configurations on the OVSD dataset.
For a metric, we use the widely accepted Coverage $C$ and Overflow $O$ \cite{coverageoverflow}, with a single value $F$-score for assessing the quality of division as the harmonic mean between $C$ and $1-O$.

Formally, as in \cite{lb_measure}, we denote $s_1,s_2,\ldots,s_m$ as the series of detected scenes, and $\tilde{s_1},\tilde{s_2},\ldots,\tilde{s_n}$ as the series of ground truth scenes, where each element $s$ is a set of shots.
The coverage $C_t$ of ground truth scene $\tilde{s}_t$ is computed as:
\begin{equation}
    C_t = \frac{ \max_{i=1, \ldots ,m}\#(s_i \cap \tilde{s}_t)}{\#\tilde{s}_t},
\end{equation}
where $\#(s)$ is the number of shots in scene $s$.
Essentially, this is the relative amount of the ground truth scene that was allocated to a single scene in the proposed division.
The overflow $O_t$ for ground truth scene $\tilde{s}_t$ is computed as:
\begin{equation}
    O_t = \frac{ \sum_{i=1}^m \left[ \#(s_i \setminus \tilde{s}_t) \cdot \min ( 1, \# (s_i \cap \tilde{s}_t ) ) \right]}{\#(\tilde{s}_{t-1})+\#(\tilde{s}_{t+1})}.
\end{equation}
Essentially, $\min ( 1, \# (s_i \cap \tilde{s}_t ) )$ is a binary indicator whether scene $s_i$ shares at least one shot with $\tilde{s}_t$, and $\#(s_i \setminus \tilde{s}_t)$ are the shots of these scenes which are not part of $\tilde{s}_t$.
Therefore this measures how much the overlapping proposed scenes extend beyond the ground truth scene normalized by the number of shots in the neighboring scenes.

These measures for each ground truth scene are aggregated into video-wide metrics as the weighted average:
\begin{equation}
    \begin{matrix} C = \sum_{t=1}^n C_t \cdot\frac{\#(\tilde{s}_t)}{\sum_i \# (\tilde{s}_i)}, & &&& O = \sum_{t=1}^n O_t \cdot\frac{\#(\tilde{s}_t)}{\sum_i \# (\tilde{s}_i)}. \end{matrix}
\end{equation}
Finally, as a single score for the quality of the scene detection, we compute the harmonic mean:
\begin{equation}
    F = 2 \cdot \frac{C \cdot (1-O)}{C + (1-O)}.
\end{equation}

\section{Additional $D$ Examples}

In Table \ref{tab:dexamples17} we present various stages of $D$ from visual features of the video La Chute D`une Plume from OVSD with the accompanied ground truth $D^*$, and in Table \ref{tab:dexamples15} the same for the video Big Buck Bunny.
In Tables \ref{tab:dexamples17epoch} and \ref{tab:dexamples15epoch} we show how the $D$ matrices and gradients evolve over a number of epochs for the videos La Chute D`une Plume and Big Buck Bunny respectively.

In general our observations are that OSG-Triplet manages to emphasize the small scenes better than large scenes, while OSG-Block is the reverse.
OSG-Block-Adjacent gives a good trade-off of emphasizing the immediate off-diagonal, but results in some low distances in the far off-diagonal.
In practice, these shouldn't affect the OSG algorithm if the intervening distances are large enough.
OSG-Prob converges more slowly, learns from the boundary edges, and gives a good trade-off as well.

\begin{figure*}[h]
\begin{center}
   \includegraphics[width=0.8\linewidth]{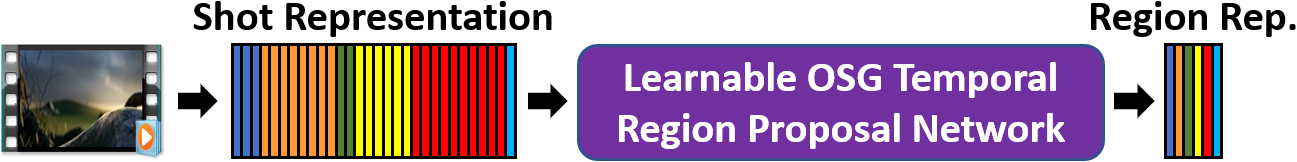}
\end{center}
   \caption{OSG-Prob as a plug-and-play temporal region proposal network.}
\label{fig:pap}
\end{figure*}

Regarding this last point, part of our motivation for OSG-Prob is to have a configuration which is specifically reliant on division locations as opposed to the block-diagonal.
Such a structure would allow OSG to be integrated into a larger learning pipeline.
For example, there are other temporal analysis tasks where division is only a part of the process.
In the weakly-supervised regime there might not be ground truth divisions with which to perform OSG-Triplet or OSG-Block.
OSG-Prob on the other hand, could be configured to perform backpropagation on a loss which reflects on the locations of division, and is inferred back to the distance values.
In this respect, our continued research involves having this component as a plug-and-play module for other tasks which can act as temporal region proposal networks (see Figure \ref{fig:pap}).

\begin{table*}[h]
    \caption{An example $D$ from the video La Chute D`une Plume from OVSD. On the left: Ground Truth ($D^*$), Orig (without an applied embedding), Epoch 0 (embedding before learning). On the right, trained examples after 20 epochs for: OSG-Triplet, OSG-Block, OSG-Block-Adjacent, and OSG-Prob, with corresponding gradients (bottom row)}
    \centering
    \begin{tabular}{cc|}
        \multicolumn{2}{c|}{Ground}  \\
        \multicolumn{2}{c|}{Truth}  \\
        \multicolumn{2}{c|}{\raisebox{-.5\height}{\includegraphics[width=0.13\linewidth]{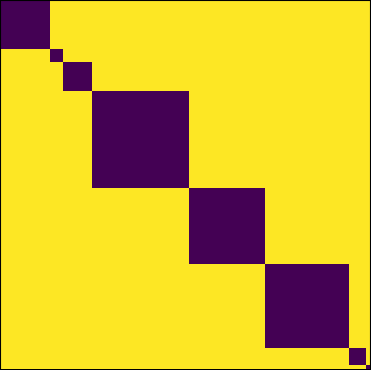}}} \\
        Orig &  Epoch 0 \\
        \raisebox{-.5\height}{\includegraphics[width=0.13\linewidth]{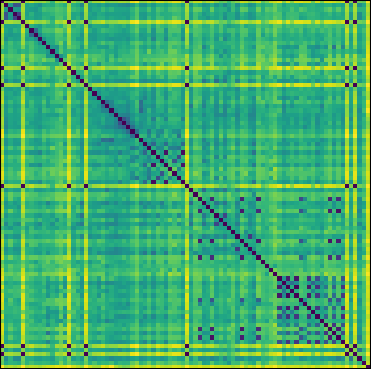}} & \raisebox{-.5\height}{\includegraphics[width=0.13\linewidth]{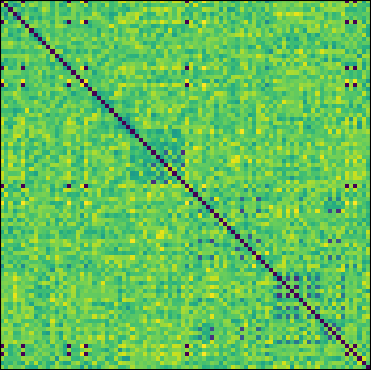}}
    \end{tabular}
    \begin{tabular}{|ccccc}
         & & OSG-Block- & \\
        OSG-Triplet & OSG-Block & Adjacent & OSG-Prob \\
        \raisebox{-.5\height}{\includegraphics[width=0.13\linewidth]{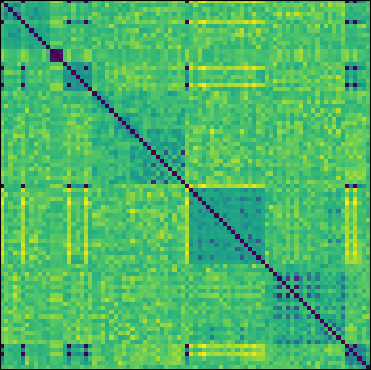}} & \raisebox{-.5\height}{\includegraphics[width=0.13\linewidth]{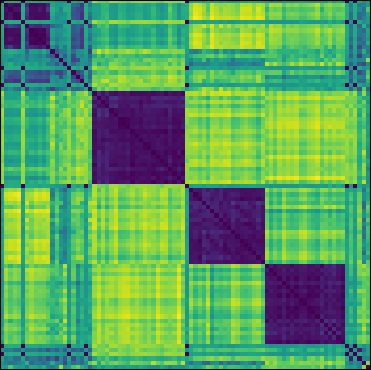}} & \raisebox{-.5\height}{\includegraphics[width=0.13\linewidth]{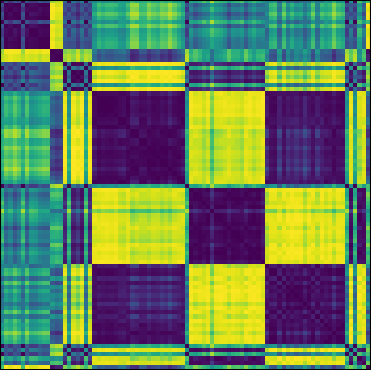}} &
        \raisebox{-.5\height}{\includegraphics[width=0.13\linewidth]{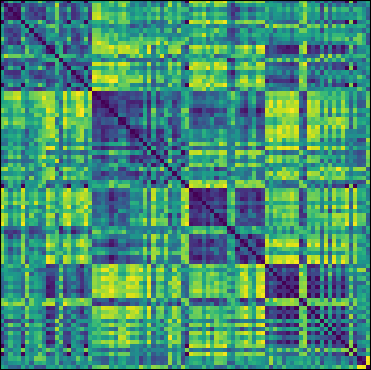}} \\
        Grad & Grad & Grad & Grad \\ \raisebox{-.5\height}{\includegraphics[width=0.13\linewidth]{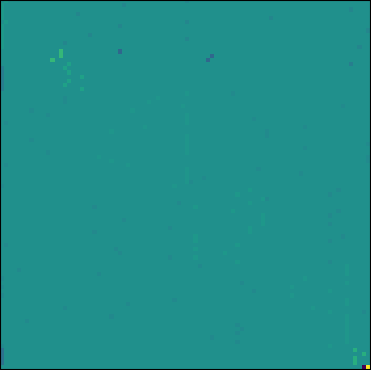}} & \raisebox{-.5\height}{\includegraphics[width=0.13\linewidth]{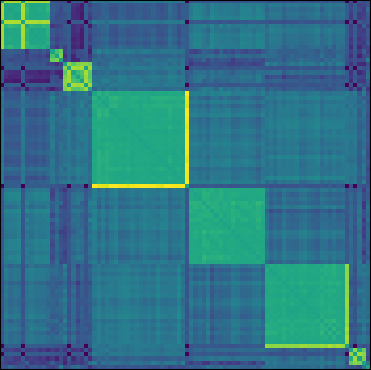}} & \raisebox{-.5\height}{\includegraphics[width=0.13\linewidth]{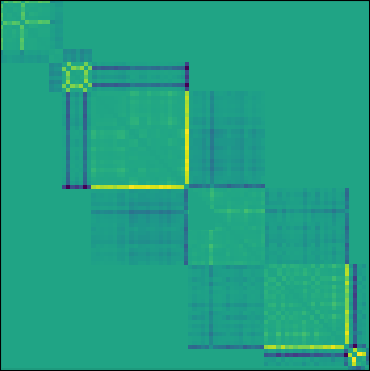}} &
        \raisebox{-.5\height}{\includegraphics[width=0.13\linewidth]{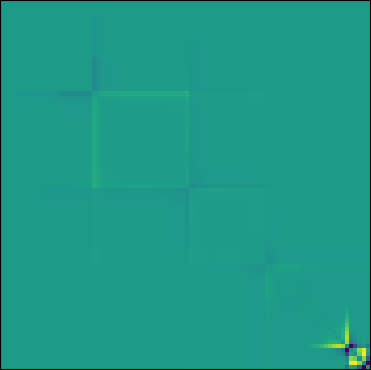}}
    \end{tabular}
    \label{tab:dexamples17}
\end{table*}

\begin{table*}[h]
    \caption{An example $D$ from the video Big Buck Bunny from OVSD. On the left: Ground Truth ($D^*$), Orig (without an applied embedding), Epoch 0 (embedding before learning). On the right, trained examples after 20 epochs for: OSG-Triplet, OSG-Block, OSG-Block-Adjacent, and OSG-Prob, with corresponding gradients (bottom row)}
    \centering
    \begin{tabular}{cc|}
        \multicolumn{2}{c|}{Ground}  \\
        \multicolumn{2}{c|}{Truth}  \\
        \multicolumn{2}{c|}{\raisebox{-.5\height}{\includegraphics[width=0.13\linewidth]{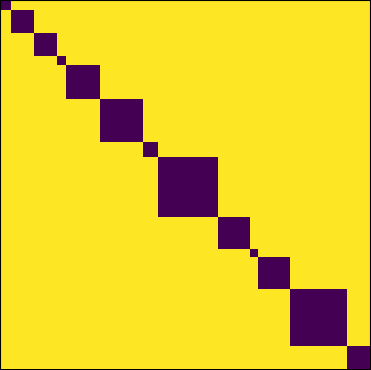}}} \\
        Orig &  Epoch 0 \\
        \raisebox{-.5\height}{\includegraphics[width=0.13\linewidth]{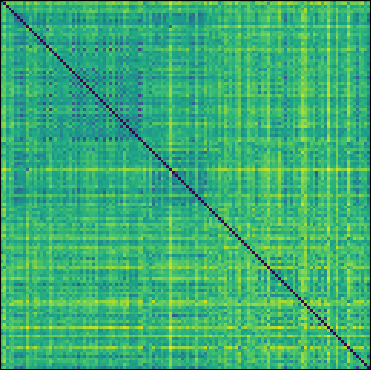}} & \raisebox{-.5\height}{\includegraphics[width=0.13\linewidth]{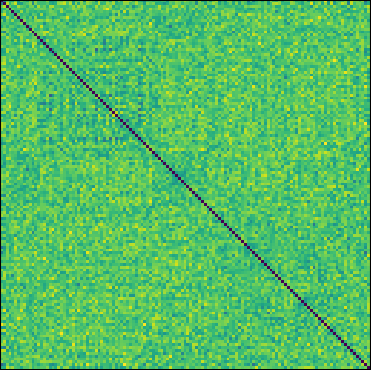}}
    \end{tabular}
    \begin{tabular}{|ccccc}
         & & OSG-Block- & \\
        OSG-Triplet & OSG-Block & Adjacent & OSG-Prob \\
        \raisebox{-.5\height}{\includegraphics[width=0.13\linewidth]{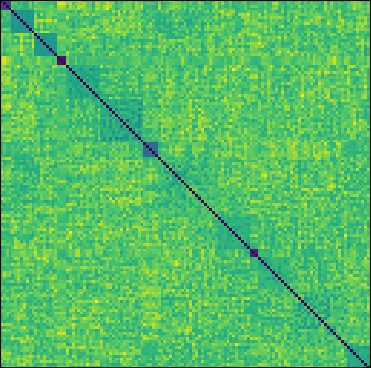}} & \raisebox{-.5\height}{\includegraphics[width=0.13\linewidth]{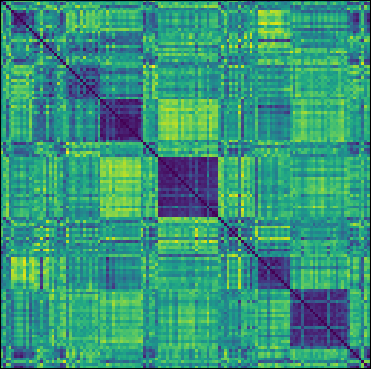}} & \raisebox{-.5\height}{\includegraphics[width=0.13\linewidth]{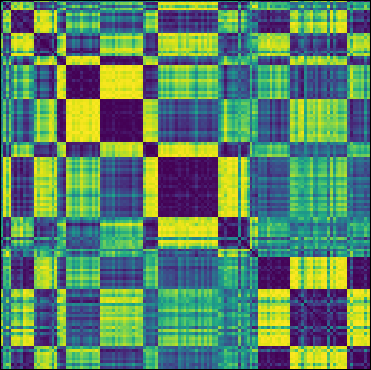}} &
        \raisebox{-.5\height}{\includegraphics[width=0.13\linewidth]{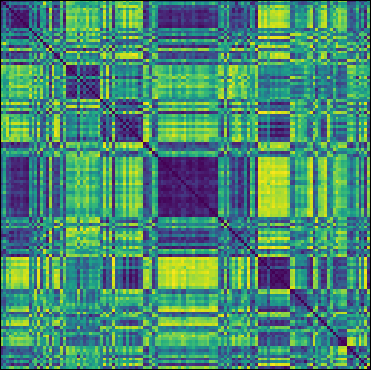}} \\
        Grad & Grad & Grad & Grad \\ \raisebox{-.5\height}{\includegraphics[width=0.13\linewidth]{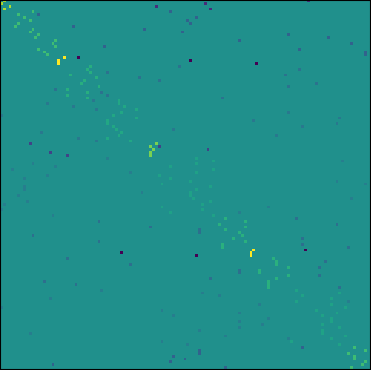}} & \raisebox{-.5\height}{\includegraphics[width=0.13\linewidth]{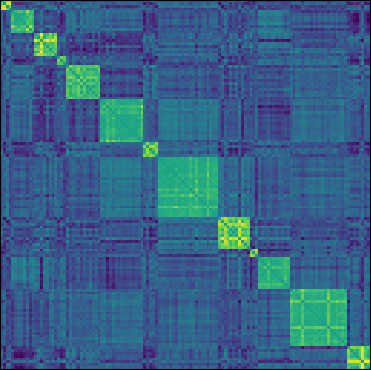}} & \raisebox{-.5\height}{\includegraphics[width=0.13\linewidth]{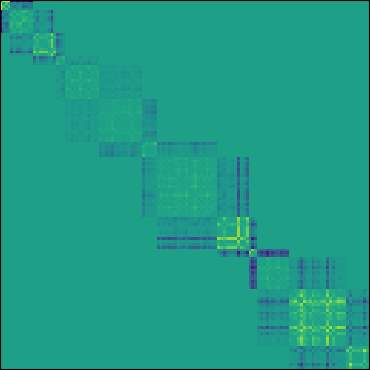}} &
        \raisebox{-.5\height}{\includegraphics[width=0.13\linewidth]{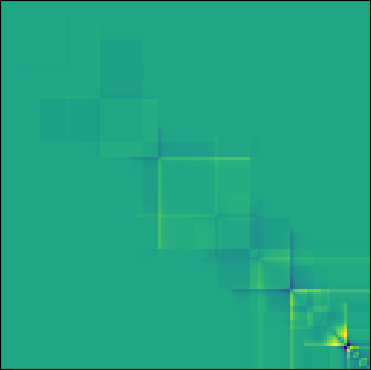}}
    \end{tabular}
    \label{tab:dexamples15}
\end{table*}

\begin{table*}[h]
    \caption{$D$ and gradients from the video La Chute D`une Plume from OVSD evolving over a number of Epochs}
    \centering
    \begin{tabular}{cccccc}
        & Epoch 0 & Epoch 5 & Epoch 10 & Epoch 15 & Epoch 20\\
        OSG-Triplet &
        \raisebox{-.5\height}{\includegraphics[width=0.13\linewidth]{Supp/17_T_D_0.png}} &
        \raisebox{-.5\height}{\includegraphics[width=0.13\linewidth]{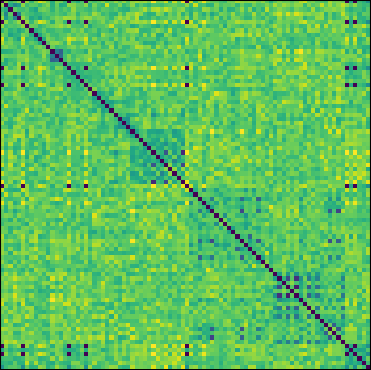}} &
        \raisebox{-.5\height}{\includegraphics[width=0.13\linewidth]{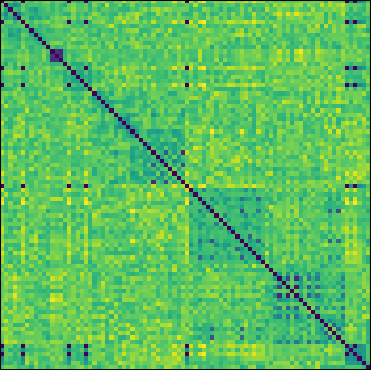}} &
        \raisebox{-.5\height}{\includegraphics[width=0.13\linewidth]{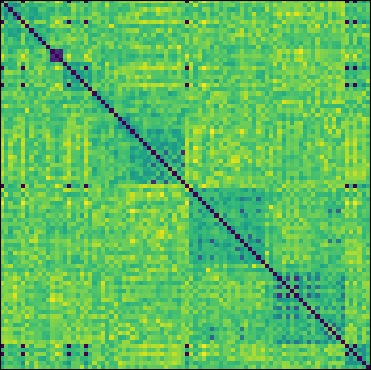}} &
        \raisebox{-.5\height}{\includegraphics[width=0.13\linewidth]{Supp/17_T_D_20.png}} \\
        \\[-1em]
        OSG-Block &
        \raisebox{-.5\height}{\includegraphics[width=0.13\linewidth]{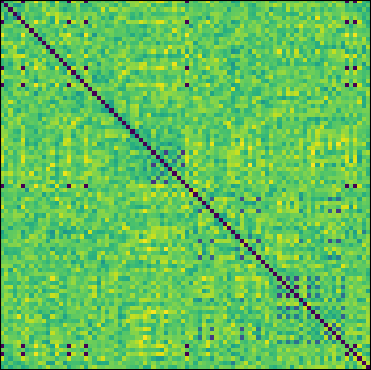}} &
        \raisebox{-.5\height}{\includegraphics[width=0.13\linewidth]{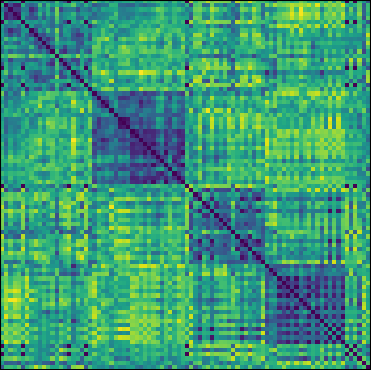}} &
        \raisebox{-.5\height}{\includegraphics[width=0.13\linewidth]{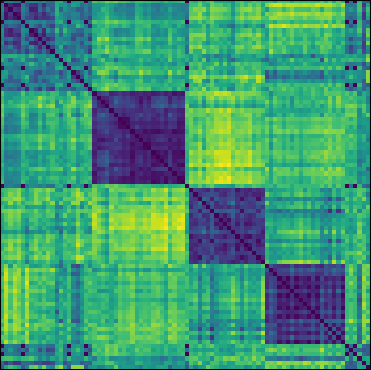}} &
        \raisebox{-.5\height}{\includegraphics[width=0.13\linewidth]{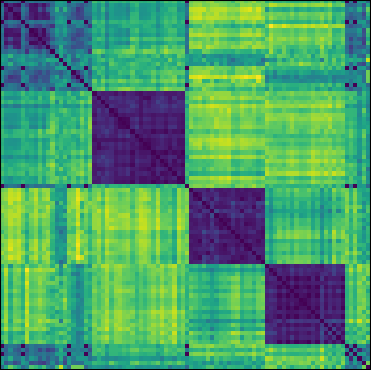}} &
        \raisebox{-.5\height}{\includegraphics[width=0.13\linewidth]{Supp/17_D_D_20.png}} \\
        \\[-1em]
        \begin{tabular}[x]{@{}c@{}}OSG-Block-\\Adjacent\end{tabular} &
        \raisebox{-.5\height}{\includegraphics[width=0.13\linewidth]{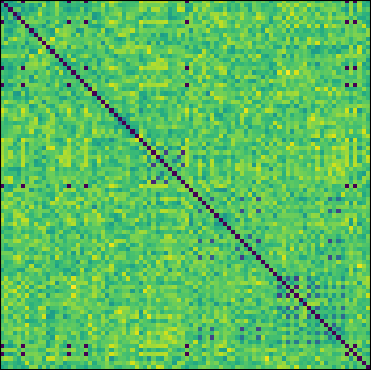}} &
        \raisebox{-.5\height}{\includegraphics[width=0.13\linewidth]{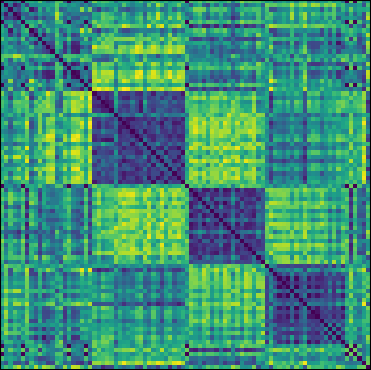}} &
        \raisebox{-.5\height}{\includegraphics[width=0.13\linewidth]{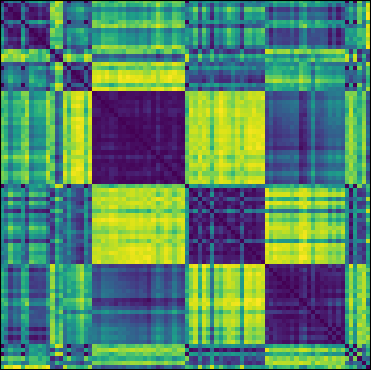}} &
        \raisebox{-.5\height}{\includegraphics[width=0.13\linewidth]{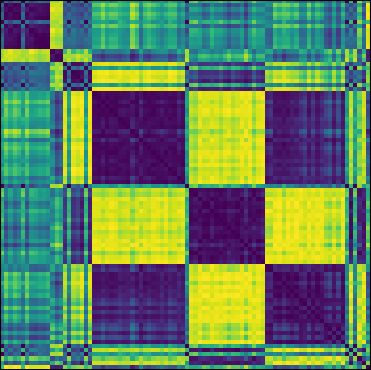}} &
        \raisebox{-.5\height}{\includegraphics[width=0.13\linewidth]{Supp/17_DA_D_20.png}} \\
        \\[-1em]
        OSG-Prob &
        \raisebox{-.5\height}{\includegraphics[width=0.13\linewidth]{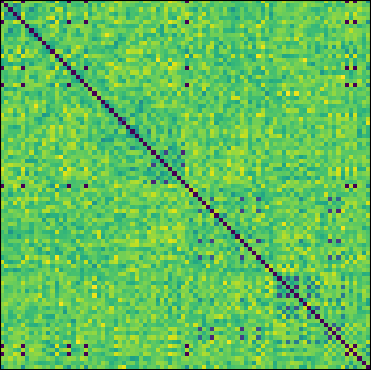}} &
        \raisebox{-.5\height}{\includegraphics[width=0.13\linewidth]{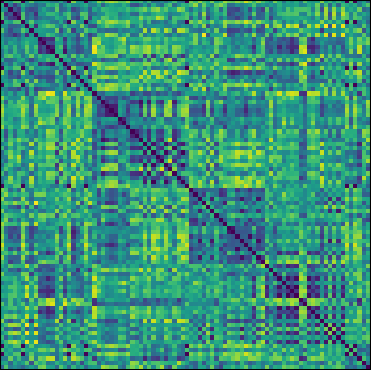}} &
        \raisebox{-.5\height}{\includegraphics[width=0.13\linewidth]{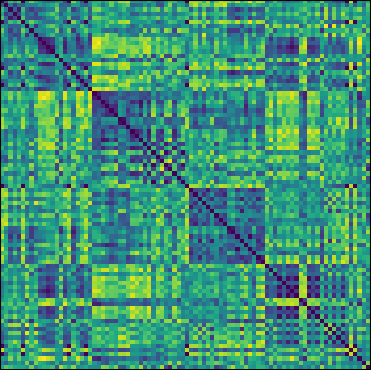}} &
        \raisebox{-.5\height}{\includegraphics[width=0.13\linewidth]{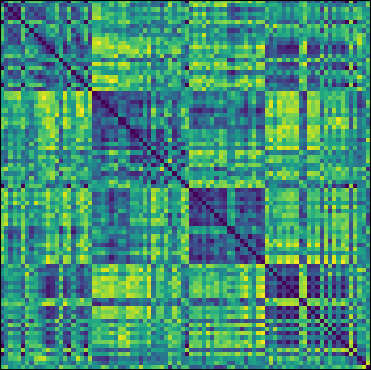}} &
        \raisebox{-.5\height}{\includegraphics[width=0.13\linewidth]{Supp/17_C_D_20.png}} \\
        \\[-1em]
        \hline
        \hline
        \multicolumn{6}{c}{Gradients} \\
        & Epoch 0 & Epoch 5 & Epoch 10 & Epoch 15 & Epoch 20\\
        OSG-Triplet &
        \raisebox{-.5\height}{\includegraphics[width=0.13\linewidth]{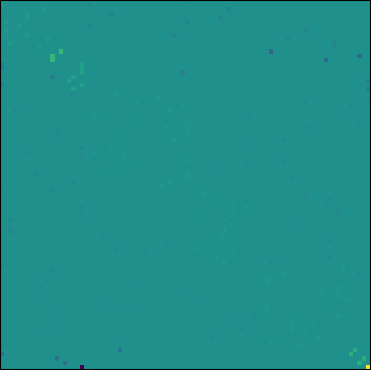}} &
        \raisebox{-.5\height}{\includegraphics[width=0.13\linewidth]{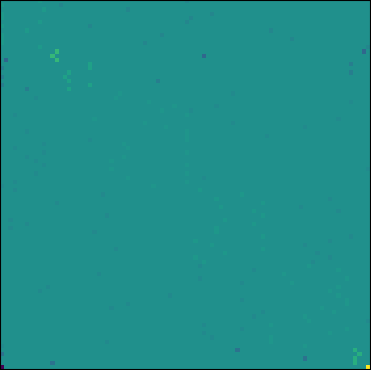}} &
        \raisebox{-.5\height}{\includegraphics[width=0.13\linewidth]{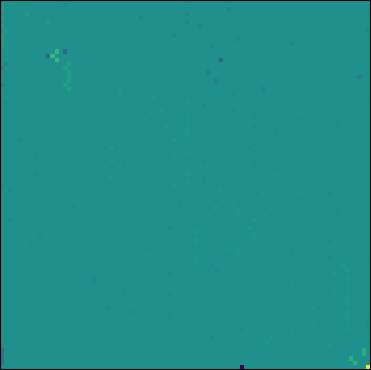}} &
        \raisebox{-.5\height}{\includegraphics[width=0.13\linewidth]{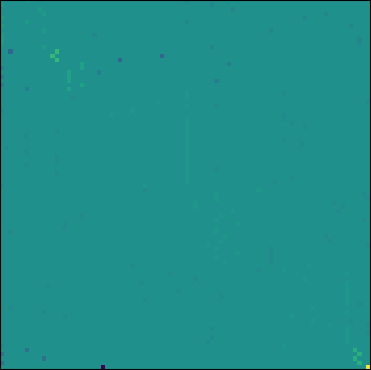}} &
        \raisebox{-.5\height}{\includegraphics[width=0.13\linewidth]{Supp/17_T_G_20.png}} \\
        \\[-1em]
        OSG-Block &
        \raisebox{-.5\height}{\includegraphics[width=0.13\linewidth]{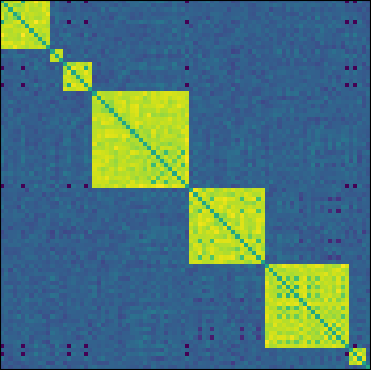}} &
        \raisebox{-.5\height}{\includegraphics[width=0.13\linewidth]{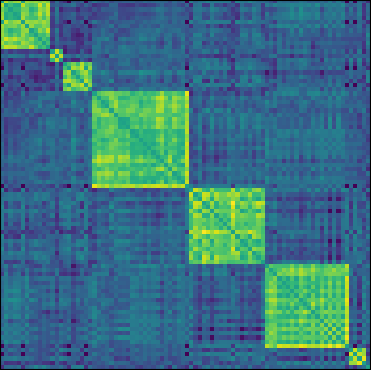}} &
        \raisebox{-.5\height}{\includegraphics[width=0.13\linewidth]{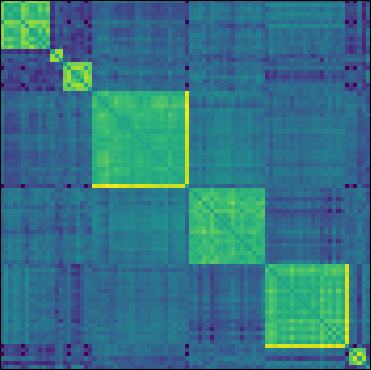}} &
        \raisebox{-.5\height}{\includegraphics[width=0.13\linewidth]{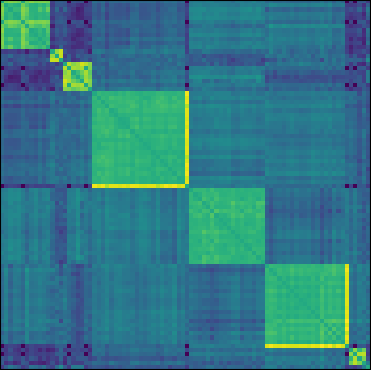}} &
        \raisebox{-.5\height}{\includegraphics[width=0.13\linewidth]{Supp/17_D_G_20.png}} \\
        \\[-1em]
        \begin{tabular}[x]{@{}c@{}}OSG-Block-\\Adjacent\end{tabular} &
        \raisebox{-.5\height}{\includegraphics[width=0.13\linewidth]{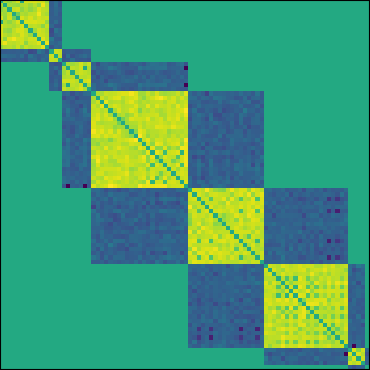}} &
        \raisebox{-.5\height}{\includegraphics[width=0.13\linewidth]{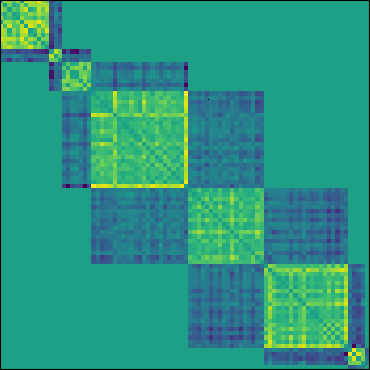}} &
        \raisebox{-.5\height}{\includegraphics[width=0.13\linewidth]{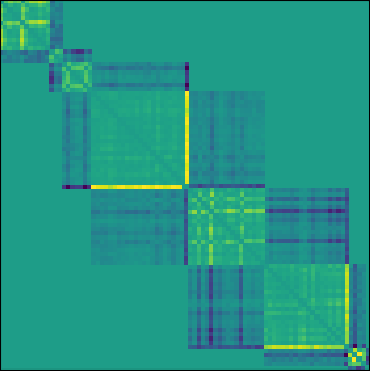}} &
        \raisebox{-.5\height}{\includegraphics[width=0.13\linewidth]{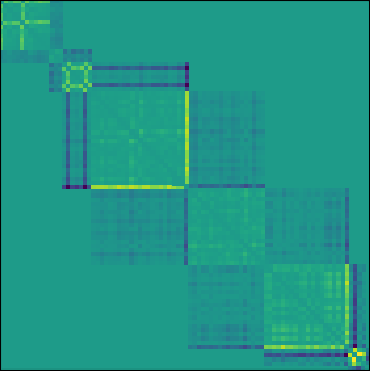}} &
        \raisebox{-.5\height}{\includegraphics[width=0.13\linewidth]{Supp/17_DA_G_20.png}} \\
        \\[-1em]
        OSG-Prob &
        \raisebox{-.5\height}{\includegraphics[width=0.13\linewidth]{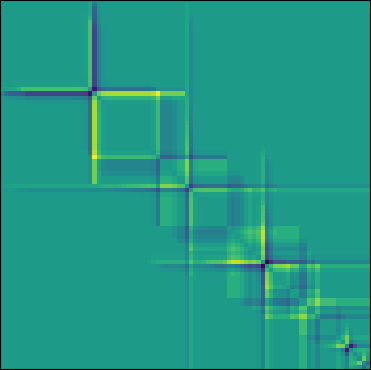}} &
        \raisebox{-.5\height}{\includegraphics[width=0.13\linewidth]{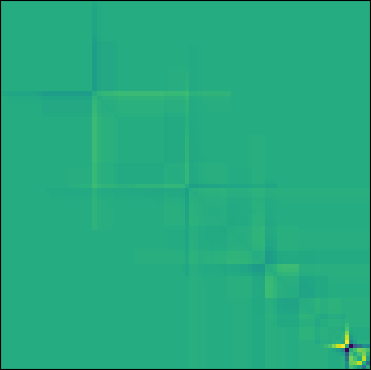}} &
        \raisebox{-.5\height}{\includegraphics[width=0.13\linewidth]{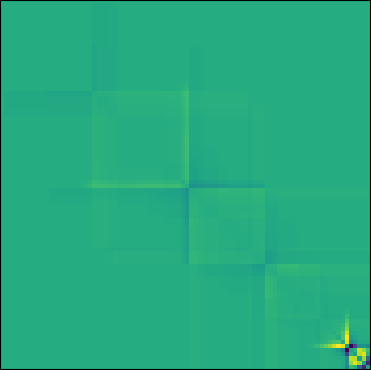}} &
        \raisebox{-.5\height}{\includegraphics[width=0.13\linewidth]{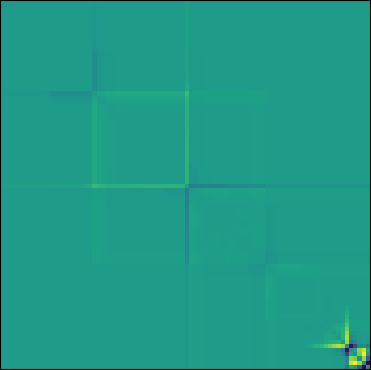}} &
        \raisebox{-.5\height}{\includegraphics[width=0.13\linewidth]{Supp/17_C_G_20.png}}
    \end{tabular}
    \label{tab:dexamples17epoch}
\end{table*}

\begin{table*}[h]
    \caption{$D$ and gradients from the video Big Buck Bunny from OVSD evolving over a number of Epochs}
    \centering
    \begin{tabular}{cccccc}
        & Epoch 0 & Epoch 5 & Epoch 10 & Epoch 15 & Epoch 20\\
        OSG-Triplet &
        \raisebox{-.5\height}{\includegraphics[width=0.13\linewidth]{Supp/15_T_D_0.png}} &
        \raisebox{-.5\height}{\includegraphics[width=0.13\linewidth]{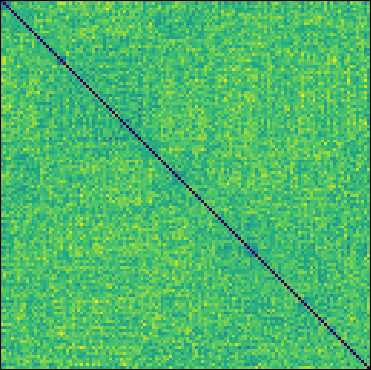}} &
        \raisebox{-.5\height}{\includegraphics[width=0.13\linewidth]{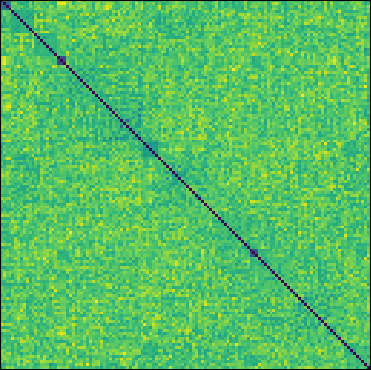}} &
        \raisebox{-.5\height}{\includegraphics[width=0.13\linewidth]{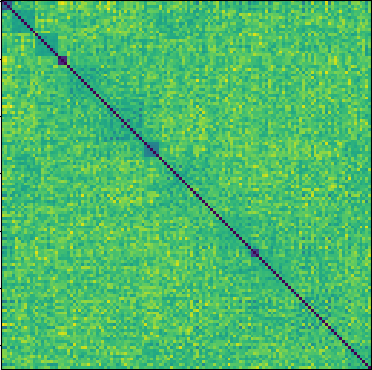}} &
        \raisebox{-.5\height}{\includegraphics[width=0.13\linewidth]{Supp/15_T_D_20.png}} \\
        \\[-1em]
        OSG-Block &
        \raisebox{-.5\height}{\includegraphics[width=0.13\linewidth]{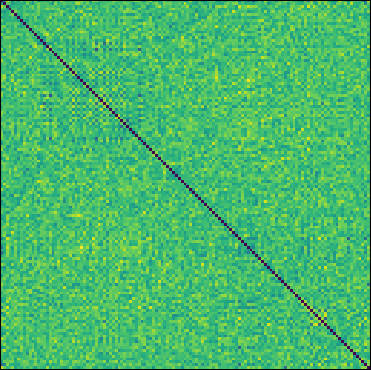}} &
        \raisebox{-.5\height}{\includegraphics[width=0.13\linewidth]{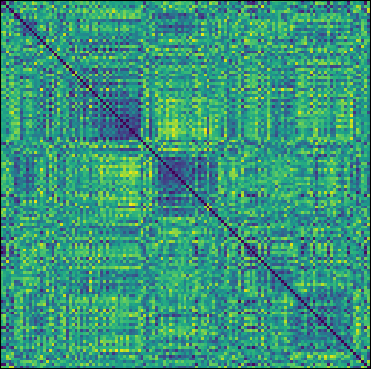}} &
        \raisebox{-.5\height}{\includegraphics[width=0.13\linewidth]{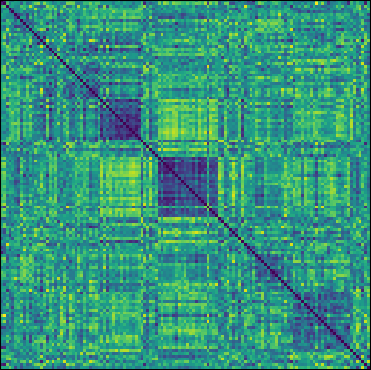}} &
        \raisebox{-.5\height}{\includegraphics[width=0.13\linewidth]{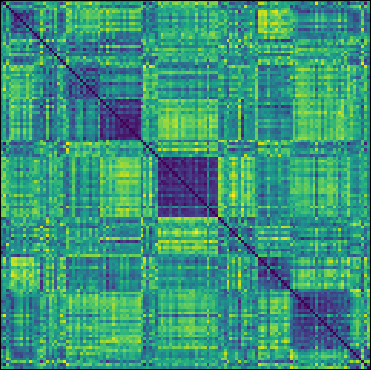}} &
        \raisebox{-.5\height}{\includegraphics[width=0.13\linewidth]{Supp/15_D_D_20.png}} \\
        \\[-1em]
        \begin{tabular}[x]{@{}c@{}}OSG-Block-\\Adjacent\end{tabular} &
        \raisebox{-.5\height}{\includegraphics[width=0.13\linewidth]{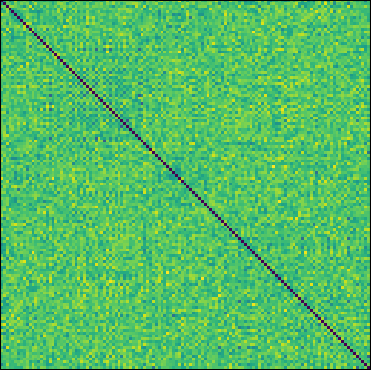}} &
        \raisebox{-.5\height}{\includegraphics[width=0.13\linewidth]{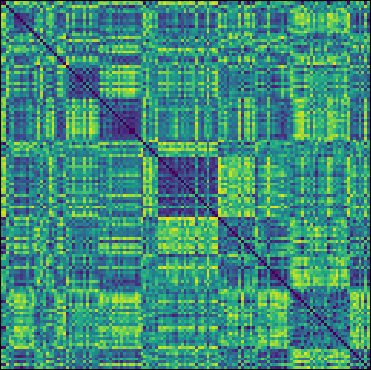}} &
        \raisebox{-.5\height}{\includegraphics[width=0.13\linewidth]{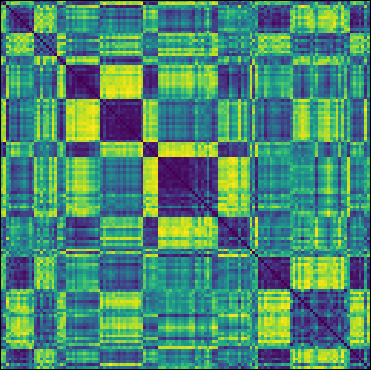}} &
        \raisebox{-.5\height}{\includegraphics[width=0.13\linewidth]{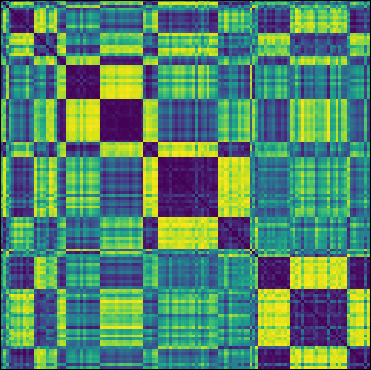}} &
        \raisebox{-.5\height}{\includegraphics[width=0.13\linewidth]{Supp/15_DA_D_20.png}} \\
        \\[-1em]
        OSG-Prob &
        \raisebox{-.5\height}{\includegraphics[width=0.13\linewidth]{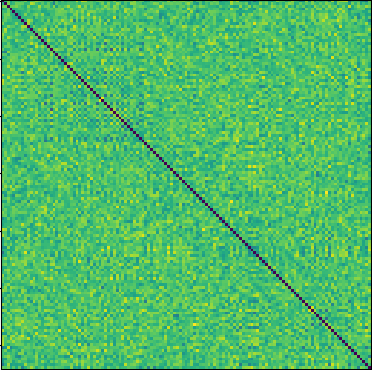}} &
        \raisebox{-.5\height}{\includegraphics[width=0.13\linewidth]{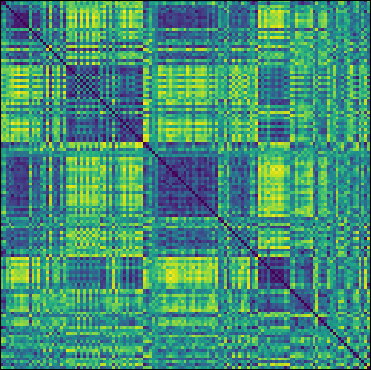}} &
        \raisebox{-.5\height}{\includegraphics[width=0.13\linewidth]{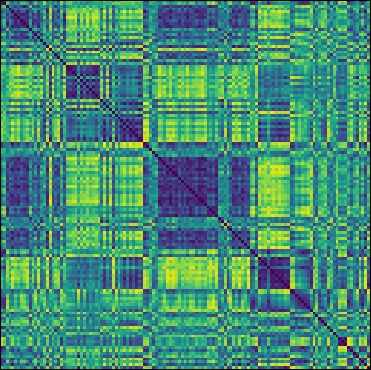}} &
        \raisebox{-.5\height}{\includegraphics[width=0.13\linewidth]{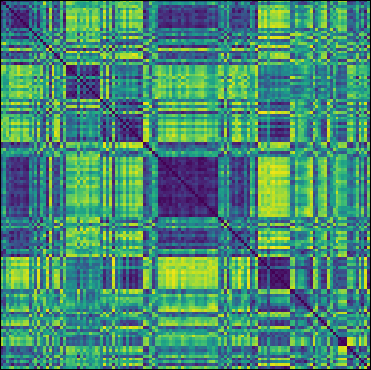}} &
        \raisebox{-.5\height}{\includegraphics[width=0.13\linewidth]{Supp/15_C_D_20.png}} \\
        \\[-1em]
        \hline
        \hline
        \multicolumn{6}{c}{Gradients} \\
        & Epoch 0 & Epoch 5 & Epoch 10 & Epoch 15 & Epoch 20\\
        OSG-Triplet &
        \raisebox{-.5\height}{\includegraphics[width=0.13\linewidth]{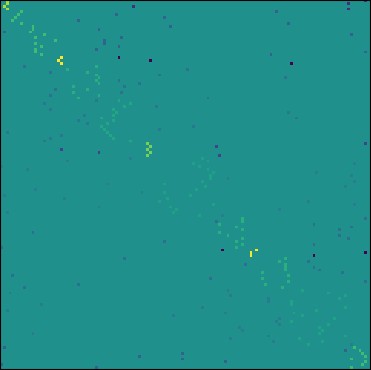}} &
        \raisebox{-.5\height}{\includegraphics[width=0.13\linewidth]{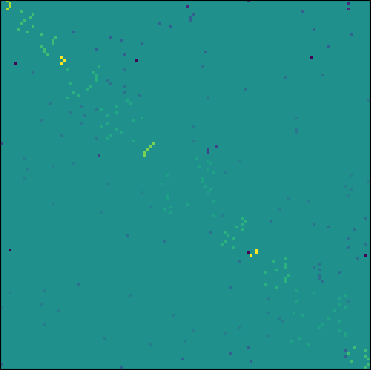}} &
        \raisebox{-.5\height}{\includegraphics[width=0.13\linewidth]{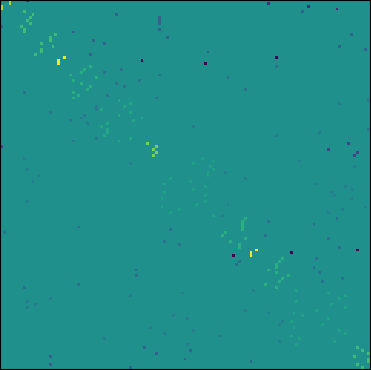}} &
        \raisebox{-.5\height}{\includegraphics[width=0.13\linewidth]{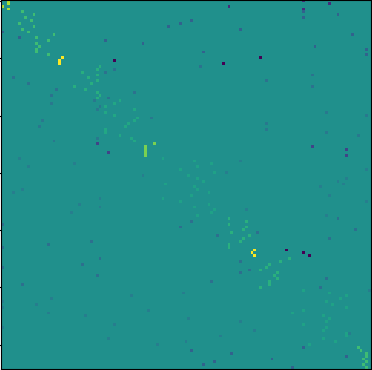}} &
        \raisebox{-.5\height}{\includegraphics[width=0.13\linewidth]{Supp/15_T_G_20.png}} \\
        \\[-1em]
        OSG-Block &
        \raisebox{-.5\height}{\includegraphics[width=0.13\linewidth]{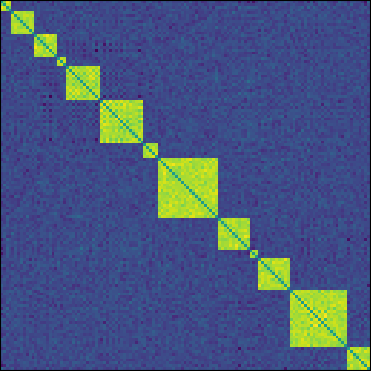}} &
        \raisebox{-.5\height}{\includegraphics[width=0.13\linewidth]{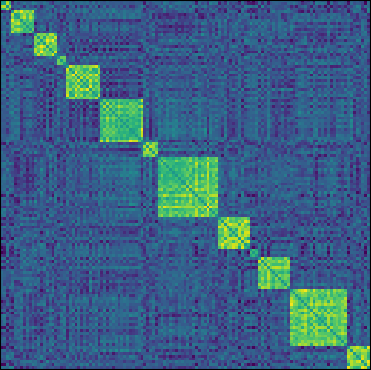}} &
        \raisebox{-.5\height}{\includegraphics[width=0.13\linewidth]{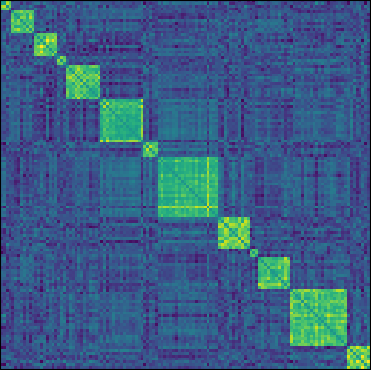}} &
        \raisebox{-.5\height}{\includegraphics[width=0.13\linewidth]{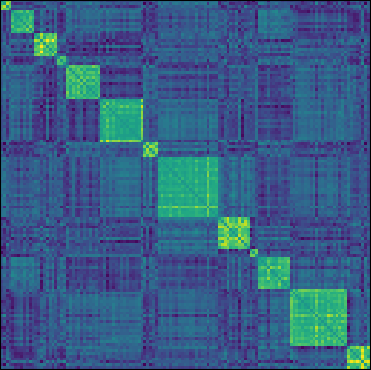}} &
        \raisebox{-.5\height}{\includegraphics[width=0.13\linewidth]{Supp/15_D_G_20.png}} \\
        \\[-1em]
        \begin{tabular}[x]{@{}c@{}}OSG-Block-\\Adjacent\end{tabular} &
        \raisebox{-.5\height}{\includegraphics[width=0.13\linewidth]{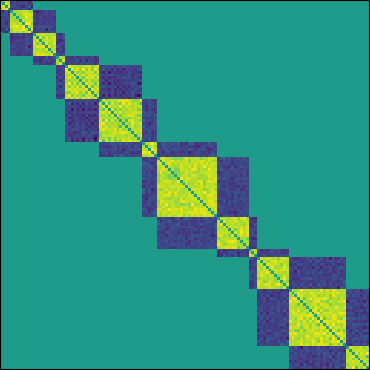}} &
        \raisebox{-.5\height}{\includegraphics[width=0.13\linewidth]{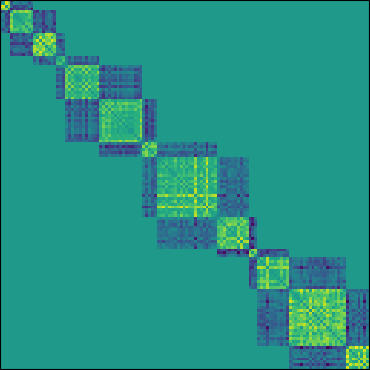}} &
        \raisebox{-.5\height}{\includegraphics[width=0.13\linewidth]{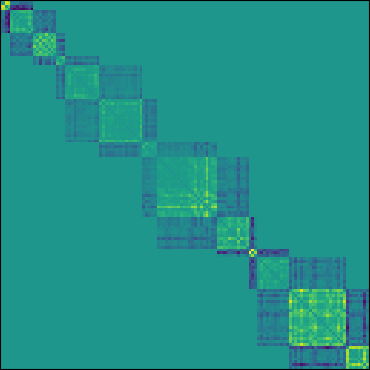}} &
        \raisebox{-.5\height}{\includegraphics[width=0.13\linewidth]{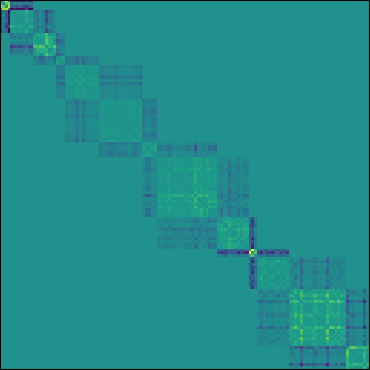}} &
        \raisebox{-.5\height}{\includegraphics[width=0.13\linewidth]{Supp/15_DA_G_20.png}} \\
        \\[-1em]
        OSG-Prob &
        \raisebox{-.5\height}{\includegraphics[width=0.13\linewidth]{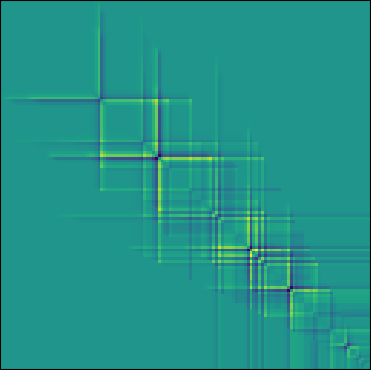}} &
        \raisebox{-.5\height}{\includegraphics[width=0.13\linewidth]{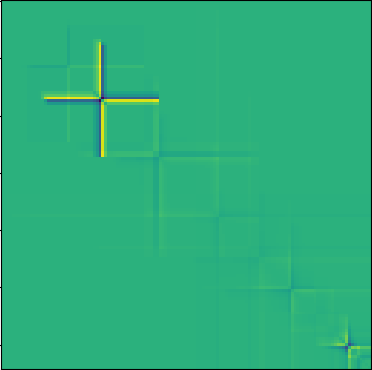}} &
        \raisebox{-.5\height}{\includegraphics[width=0.13\linewidth]{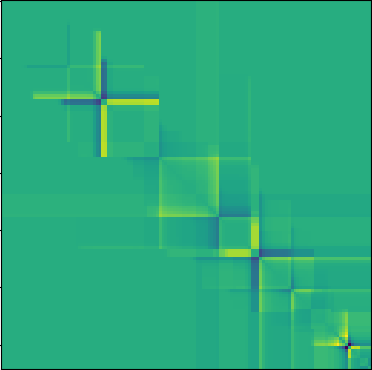}} &
        \raisebox{-.5\height}{\includegraphics[width=0.13\linewidth]{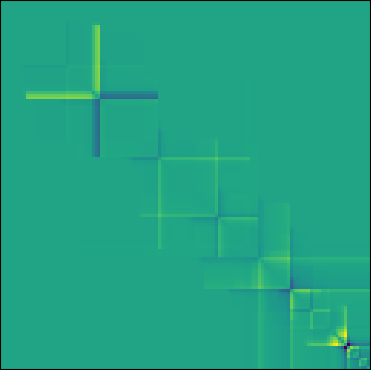}} &
        \raisebox{-.5\height}{\includegraphics[width=0.13\linewidth]{Supp/15_C_G_20.png}}
    \end{tabular}
    \label{tab:dexamples15epoch}
\end{table*}

\section{Additional $T$ Examples}

\begin{figure*}[h]
\centering
\includegraphics[width=\linewidth]{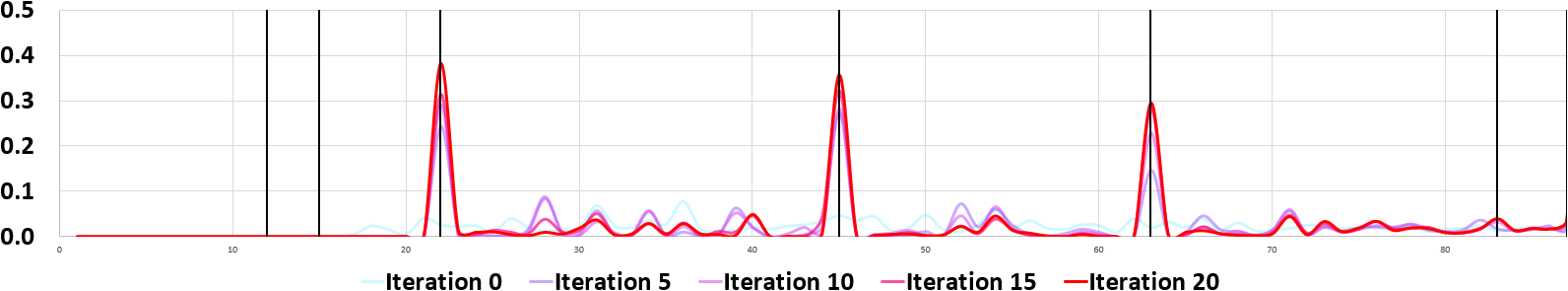}
   \caption{Progression of $T$ as a function of $i$ (shot number) over a number of iterations for video La Chute D`une Plume. Graphs go from translucent blue to opaque red as iterations progress (best viewed in color). Vertical black lines indicate ground truth divisions.}
\label{fig:t_evo_17}
\end{figure*}

\begin{figure*}[h]
\centering
\includegraphics[width=\linewidth]{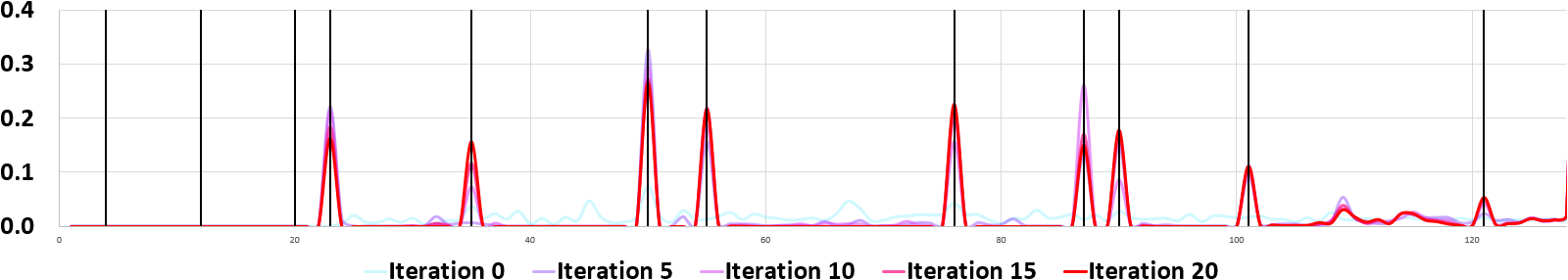}
   \caption{Progression of $T$ as a function of $i$ (shot number) over a number of iterations for video Big Buck Bunny. Graphs go from translucent blue to opaque red as iterations progress (best viewed in color). Vertical black lines indicate ground truth divisions.}
\label{fig:t_evo_15}
\end{figure*}

In Figure \ref{fig:t_evo_17} and \ref{fig:t_evo_15} we show the progression of the values of $T(i)$ over a number of iterations for videos La Chute D`une Plume and Big Buck Bunny respectively.
We can see that as the iterations progress, $T$ raises the probability at the ground truth points of division.
The probability is lowered for locations with no true division even though this is not specifically enforced by the loss but rather an outcome of the construction of $T$.

Specifically we note that in these instances the small beginning scenes proved difficult for the network to emphasize $T$ on.
We speculate that this is due to the formulation, where smaller values of $n$ inspect longer and longer sequences (see the formulation in the paper).
Possible future work could be to formulate an additional mirrored OSG which inspects the $D$ matrix backwards.

\section{Additional Visual Results}
\label{sup:visual}

\begin{figure*}[h]
\centering
\includegraphics[width=\linewidth]{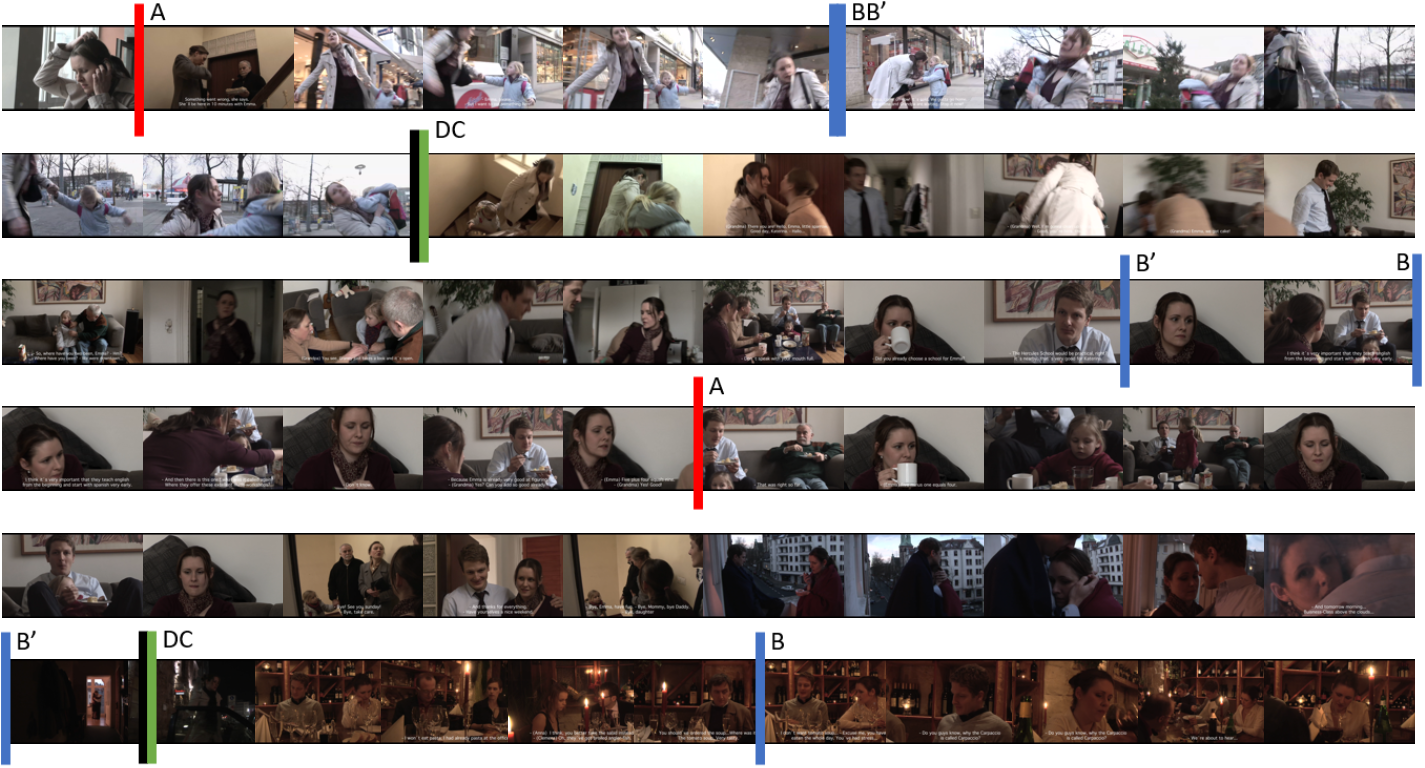}
   \caption{Qualitative results of configurations on shots 68 through 128 of the video 1000 Days from the OVSD dataset. Points of division marked by A. OSG-Triplet (red) B. OSG-Block (blue) B`. OSG-Block-Adjacent (blue) C. OSG-Prob (green) and D. Ground truth (black).}
\label{fig:frames_1000}
\end{figure*}

\begin{figure*}[h]
\centering
\includegraphics[width=\linewidth]{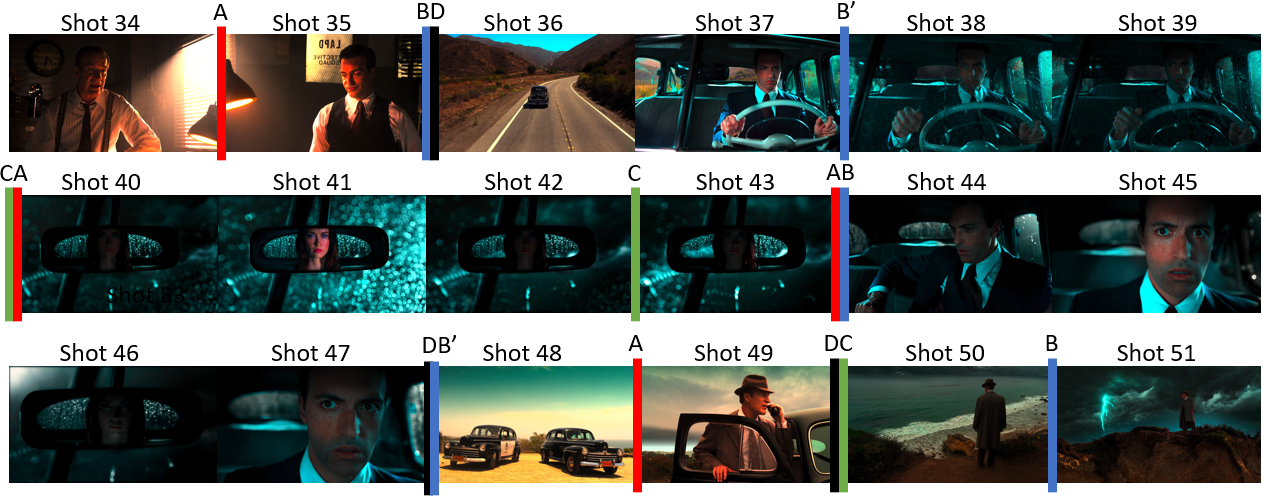}
   \caption{Qualitative results of configurations on a section of the video Meridian from the OVSD dataset. Points of division marked by A. OSG-Triplet (red) B. OSG-Block (blue) B`. OSG-Block-Adjacent (blue) C. OSG-Prob (green) and D. Ground truth (black).}
\label{fig:frames_Meridian}
\end{figure*}

\begin{figure*}[h]
\centering
\includegraphics[width=\linewidth]{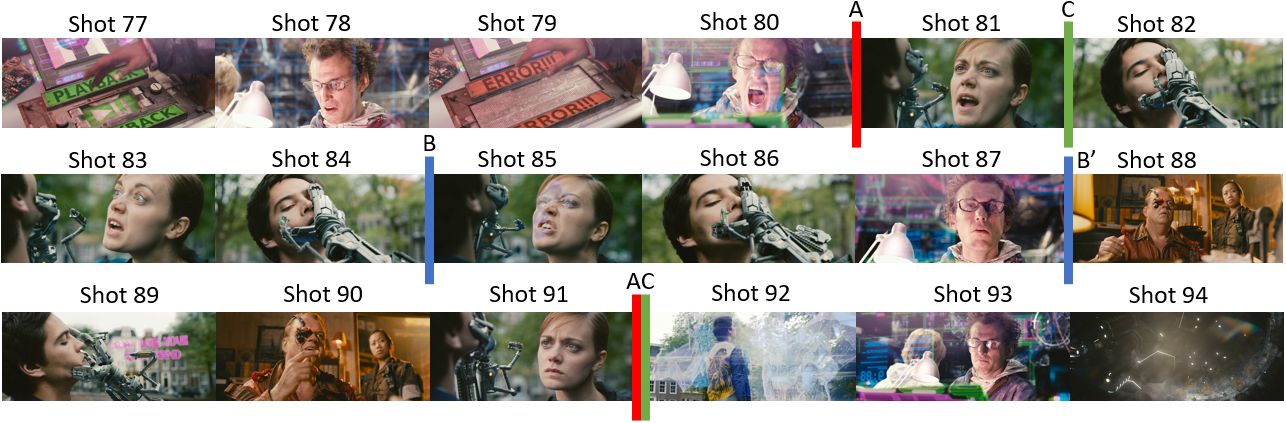}
   \caption{Qualitative results of configurations on a section of the video Tears of Steel from the OVSD dataset. Points of division marked by A. OSG-Triplet (red) B. OSG-Block (blue) B`. OSG-Block-Adjacent (blue) C. OSG-Prob (green). The shots are part of a single complex ground truth scene.}
\label{fig:frames_ToS}
\end{figure*}

Figures \ref{fig:frames_1000}, \ref{fig:frames_Meridian}, and \ref{fig:frames_ToS}, show results on sections of videos from the OVSD dataset.
In general, we can see divisions which result in reasonable and often precise scene divisions.
Using these divisions for applying video understanding and classification technologies will undoubtedly be superior over applying them on the entire video or on naive uniform divisions.
Specifically, Figure \ref{fig:frames_ToS} is a single scene from the video Tears of Steel which includes intricate character and setting changes.
This portrays the complexity of the task and the challenges that the method needs to overcome.
Despite the fact that all of the proposed scene divisions in this case are technically false, we note that they present a plausible division to story-units, and can be useful for a variety of downstream tasks.

\end{document}